\documentclass[letterpaper, 10 pt, conference]{ieeeconf}  
\let\labelindent\relax
\IEEEoverridecommandlockouts                              
\usepackage{graphicx}
\graphicspath{{Figures/}}
\DeclareGraphicsExtensions{.pdf}
\usepackage{cite}
\usepackage{algorithm,algorithmic}
\usepackage{amssymb,mathtools}
\usepackage{booktabs}
\usepackage[inline,shortlabels]{enumitem}
\usepackage[normalem]{ulem}
\usepackage{xcolor}
\usepackage{siunitx}
\usepackage{hyperref}
\usepackage{amsmath}
\usepackage{multirow}
\usepackage{titlesec}
\usepackage{flushend}  % to balance columns on last page if it's not full
\def\Plus{\texttt{+}}

\DeclareMathOperator*{\argmin}{arg\,min}

\newcommand{\eg}{e.g., }

\newcommand{\ie}{i.e., }

\newcommand{\ra}[1]{\renewcommand{\arraystretch}{#1}}

\def \figurename {Fig.}
\def \figurenamelong {Figure}

\def \Sectionname {Section}
\def \Tablename {Table}
\def \Eqname {Eq.}

\def \algoname {Algorithm}
\newcommand{\fref}[1]{\figurename~\ref{#1}}
\newcommand{\Fref}[1]{\figurenamelong~\ref{#1}}
\newcommand{\tref}[1]{\Tablename~\ref{#1}}
\newcommand{\sref}[1]{\Sectionname~\ref{#1}}

\newcommand{\eref}[1]{\Eqname~(\ref{#1})}
\newcommand{\algoref}[1]{\algoname~\ref{#1}}
\newcommand{\qref}[1]{\ref{#1}}

\newcommand{\newtext}[1]{\textcolor{black}{#1}}

\def \expname {Experiment }

\newcounter{expcounter}

\title{\LARGE \bf
Learning physics-informed simulation models for soft robotic manipulation: A case study with dielectric elastomer actuators
}
\author{Manu Lahariya$^{1*}$, Craig Innes$^{2}$, Chris Develder$^{1}$ and Subramanian Ramamoorthy$^{2}$% <-this % stops a space
\thanks{$^{1}$Authors are with IDLab, Ghent University -- imec, Technologiepark-Zwijnaarde 126, 9052~Ghent, Belgium, }%e-mail:$\{$manu.lahariya, Chris.Develder$\}$@ugent.be.}
\thanks{$^{2}$Authors are with the School of Informatics, University of Edinburgh, 10 Crichton St, EH8 9AB, United Kingdom, 
}%$\{$c.innes,s.ramamoorthy$\}$@ed.ac.uk}%
\thanks{*Corresponding Author:  e-mail:~manu.lahariya@ugent.be.
}%$\{$c.innes,s.ramamoorthy$\}$@ed.ac.uk}%
}
\begin{document}
\maketitle
\thispagestyle{empty}
\pagestyle{empty}
%%%%%%%%%%%%%%%%%%%%%%%%%%%%%%%%%%%%%%%%%%%%%%%%%%%%%%%%%%%%%%%%%%%%%%%%%%%%%%%%
\begin{abstract}
Soft actuators offer a safe, adaptable approach to tasks like gentle grasping and dexterous manipulation. Creating accurate models to control such systems however is challenging due to the complex physics of deformable materials. %Traditional particle based simulators can generate physically unrealistic results.
Accurate Finite Element Method (FEM) models incur prohibitive computational complexity for closed-loop use. Using a differentiable simulator is an attractive alternative, but their applicability to soft actuators and deformable materials remains under-explored.
%for such soft robotic tasks is a challenging task
%A controller for such actuators can be learned using a simulator of the system physics. 
%Yet, defining an accurate simulator for a manipulation task is a major challenge due to complicated physics, as it needs to model the intrinsic actuator material behaviour along with the contact/motion dynamics.
% Traditionally, particle based simulators were used, that are not guaranteed to reflect physical constraints. 
This paper presents a framework that combines the advantages of both. We learn a differentiable model consisting of a material  properties neural network and an analytical dynamics model of the remainder of the manipulation task. This physics-informed model is trained using data generated from FEM, and can be used for closed-loop control and inference. We evaluate our framework on a dielectric elastomer actuator (DEA) coin-pulling task. We simulate the task of using DEA to pull a coin along a surface with frictional contact, using FEM, and evaluate the physics-informed model for simulation, control, and inference. 
%Results show that our model provides $\le$ 5\% simulation error compared to FEM and forms the basis for an MPC controller that outperforms (iterations to convergence) a model-free actor-critic policy, a heuristic policy, and a PD controller. 
Our model attains $\le$\,5\% simulation error compared to FEM, and we use it as the basis for an MPC controller that requires fewer iterations to converge than model-free actor-critic, PD, and heuristic policies. 
%To the best of our knowledge, this is the first effort to design a framework to learn model-based control for soft robots.
%physics based model, which can be used as a differentiable simulator for any soft robotic manipulation task. 

%This model divides the system physics into two parts, the intrinsic behaviour of the deformable actuator (\eg electromechanical behaviour) and the dynamics of the object being manipulated. 

%For evaluation, we define a pulling task using a dielectric elastomer actuator. 

%We use Finite Elements Methods to generate data to learn the intrinsic behaviour of dielectric elastomer. 

% Experiments are designed to test our model closed-loop simulation and control (model predictive controller utilizing the differentiablity of our model). 

% The results show that the our physics based model provides less than 5\% error compared to FEM in closed-loop simulation, and can be utilized to learn a 70\% faster controler policy compared to a naive policy.
\end{abstract}
%%%%%%%%%%%%%%%%%%%%%%%%%%%%%%%%%%%%%%%%%%%%%%%%%%%%%%%%%%%%%%%%%%%%%%%%%%%%%%%%
%%%%%%%%%%%%%%%%%%%%%%%%%%%%%%%%%%%%%%%%%%%%%%%%%%%%%%%%%%%%%%%%%%%%%%%%%%%%%%%%
%%%%%%%%%%%%%%%%%%%%%%%%%%%%%%%%%%%%%%%%%%%%%%%%%%%%%%%%%%%%%%%%%%%%%%%%%%%%%%%%
\section{INTRODUCTION}
\label{sec:introduction}
%%%%%%%%%%%%%%%%%%%%%%%%%%%%%%%%%%%%%%%%%%%%%%%%%%%%%%%%%%%%%%%%%%%%%%%%%%%%%%%%
%%%%%%%%%%%%%%%%%%%%%%%%%%%%%%%%%%%%%%%%%%%%%%%%%%%%%%%%%%%%%%%%%%%%%%%%%%%%%%%%

Soft robotic actuators provide a safe, adaptive, low-cost solution for movement tasks such as grasping and dexterous manipulation~\cite{KIM_2013_softrobots}. Precision manipulation using soft actuators however is a major challenge, as it requires modelling the deformable actuators' dynamic within the context of the manipulation task~\cite{Rus_2015_softrobotics}. Such models are then used to learn accurate control strategies via simulation~\cite{Yin_2021_ModellingDeformable}. Recently \emph{differentiable simulators} have been used to learn controllers in closed-loop scenarios by allowing the use of gradient-based optimization methods  (\eg Model Predictive Control, MPC)~\cite{Atab_2020_SoftReview}. They have also been used for inference and data generation tasks.

Simulating deformable robots and contact rich manipulation is expensive~\cite{Yin_2021_ModellingDeformable}. Traditional methods model such dynamics by decomposing their geometry. For example, \emph{Position Based Dynamics} approximates multi-body physics by deconstructing the system into particles\cite{Todorov_2012_mujoco}.
However, these methods fail to accurately capture the true underlying physics, making it difficult to meaningfully interpret or constrain the particles. The continuum mechanics and contact dynamics of deformable materials are difficult to model with such approximate methods, leading to physically unrealistic results, which hinders model-based control. Physically accurate simulation of soft robotic manipulation requires modelling the underlying equations, defined by complex Ordinary/Partial Differential Equations (ODEs/PDEs).

Finite Element Methods (FEMs) provide a numerical method for solving such equations. Yet despite the ability of FEMs to accurately model such phenomena, integrating FEM simulation with closed-loop control is challenging due to their computationally expensive meshing: unless the meshes are dense and cover the domain, fidelity is poor.

\begin{figure}[t!]
    \centering
    \includegraphics[width=0.38\textwidth]{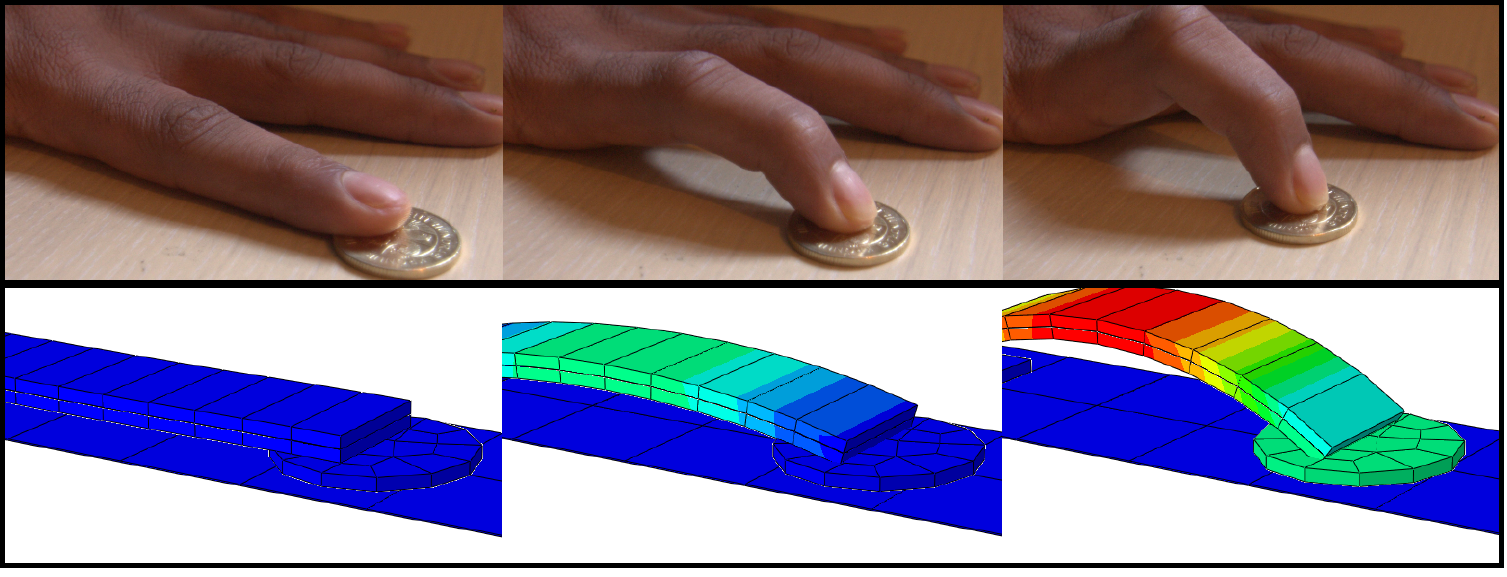}
    \caption{Coin pulled by a dielectric elastomer actuator (DEA)---a soft actuator that deforms under electric actuation. Our framework learns accurate physics-informed differentiable simulators and model-based control for such soft robot manipulation.}
    \label{fig:intro-coin}
\end{figure}

This paper's key idea is to generate data from an accurate (but slow) FEM model to learn an approximate (but fast) physics-informed model $f$ for soft robotic manipulation. Our framework uses $f$ as a differentiable simulator for simultaneous closed-loop control and inference. Our model $f$ is composed of two parts: a \emph{material network} $m$ --- which approximates deformable material behaviour (\eg hyperelastic) --- and \emph{dynamics} $d$ --- equations representing the physical context of manipulation task (sliding motion under frictional contact with the surface).

%We apply our framework to a soft robotic pulling task using Dielectric Elastomer Actuators (DEAs). 

\newtext{Our framework's objective is to learn fast physics-informed models that can be evaluated in real-time without significant loss in accuracy, and to use these models for control. As an illustrative use case, we focus on soft robotic pulling task using Dielectric Elastomer Actuators (DEAs).} DEAs are soft actuators  made using electroactive polymers that convert electrical work to mechanical work via expanding or bending motion. In our task, the goal is to pull a stationary coin by deforming the free end of the DEA (\fref{fig:intro-coin}). We learn the physics-informed model for this pulling motion ${f}$, and evaluate its accuracy as a simulator against FEM simulations. For control, we use the differentiability of $f$ to learn a model-based control policy (with the solver from~\cite{Brandon_2019_MPCDiff}) and infer the parameters of the system's dynamics.\footnote{For the case study of DEA pulling, coin mass $m_c$ and kinetic friction coefficient $\mu_b$ are inferred. For details, refer to~\sref{sec:SetupD}.}

Our main contributions are: 
\begin{enumerate*}[(i)]
    \item \label{c:framework} a closed-loop control framework for soft robotic manipulation, that uses a differentiable physics-informed model $f$ trained using FEM (\sref{sec:framework}),
    \item \label{c:FEMmodel} the design of an exemplary DEA pulling task (\sref{sec:roboticpullingsetup}), that is simulated in FEM (\sref{sec:FEADEA}), and
    \item \label{c:evaluation} 
    performance evaluation of model $f$ and its use in closed-loop control. 
\end{enumerate*}
For the latter, we compare simulation accuracy of $f$ with both a FEM model and a baseline neural network. Additionally, we compare the model-based control policy (MPC with $f$), with
\begin{enumerate*}[(i)]
\item a model-free control policy (learnt using the Soft Actor-Critic SAC algorithm~\cite{haarnoja_2018_SAC}),
\item a PD control policy (evaluated previously for DEA control~\cite{Karner_2021_DEAPIDcontrol}), and
\item a heuristic control policy\footnote{The heuristic policy linearly ramps up actuation voltage until the `episodic' task terminates. For details, refer to~\sref{sec:experimentalresults}} (inspired by typical soft-robotic control policies~\cite{Yin_2021_ModellingDeformable}).
\end{enumerate*}
We design experiments (\sref{sec:experimentalresults}) across 8 DEA pulling setups to evaluate our framework and answer the following questions:       
\begin{enumerate}[label=\textbf{(Q\arabic*)},ref={Q\arabic*},topsep=0pt,leftmargin=*,itemsep=0pt,labelsep=6pt]    
    \item \label{q:simulation} How to define $f$ using the physical laws of the system? What is the simulation accuracy of $f$ in a system with new \emph{unknown} parameters (\eg frictional coefficient)?
    \item \label{q:control} What is the performance of model-based control policy (based on $f$), compared to other control policies?
    \item \label{q:inference} What is the accuracy of the inferred model parameters?
    % Evaluation of the trained physics-informed model $f$ for closed-loop simulation and control of the DEA coin pulling.
\end{enumerate}

%The physics-informed model provides $\le$5\% closed-loop simulation error and its differentiability can be used to learn a controller. 
%It can be  trained for any soft robotic manipulation task (\sref{sec:framework}). 
%We define a DEA pulling task, simulate the task using FEM modelling and derive its dynamics (\sref{sec:DEA} and~\sref{sec:FEM}).

Our results show $f$ provides $\le$\,5\% simulation error compared to FEM. Further, in closed-loop control, an MPC using $f$ outperforms all other policies, while simultaneously inferring system properties with $\le$\,10\% error. Vidoes of the manipulation task and other supplementary materials are available at \href{https://sites.google.com/view/phy-informed-sim-soft-robot/home}{https://sites.google.com/view/phy-informed-sim-soft-robot/home}.

%%%%%%%%%%%%%%%%%%%%%%%%%%%%%%%%%%%%%%%%%%%%%%%%%%%%%%%%%%%%%%%%%%%%%%%%%%%%%%%%
\subsection{Related Work}
\label{sec:relatedwork}
%%%%%%%%%%%%%%%%%%%%%%%%%%%%%%%%%%%%%%%%%%%%%%%%%%%%%%%%%%%%%%%%%%%%%%%%%%%%%%%%

Soft robots are inspired by biological systems, \eg where animals use muscles to achieve safe actuation and control~\cite{KIM_2013_softrobots}. Engineers use soft actuators to develop similarly safe, quick, adaptable, and precise robotic manipulation~\cite{Rus_2015_softrobotics}. These soft actuators generate mechanical work under a specific actuation, \eg shape memory alloys respond to thermal actuation, hydraulic actuators respond to pressure, etc. Learning control for soft actuators requires accurate simulation models that are used inside the control loop~\cite{Atab_2020_SoftReview}. Designing simulator models for soft actuators is a challenging task, traditionally using particle based models (\eg liquids~\cite{tataia_2017_pouring}). In recent years, researchers are using FEM modelling that allows highly accurate modelling of deformable materials (\eg fabric~\cite{Coevoet_2019_cloth}, composite materials~\cite{Muzel_2020_FEMcomposite}).

Dielectric elastomers (DE) are electroactive polymers that produce deformation under the influence of an external electric field. DEA are soft actuators that use thin layers of DE materials to achieve actuation under the stimulus of electric activation. DEAs provide fast and large deformation, are lightweight, and have a high energy density, which makes them promising candidates for soft robotic applications~\cite{Gupta_2019_DEAreview}. Hence, DEA has been explored to design soft robotic grippers~\cite{Zhou_2020_DEAFEM},
underwater robots~\cite{Shintake_2016_DEAUnderwater}, crawling robots~\cite{DuDuta_2020_multimodel_locomotion}, etc.

\newtext{There have been several previous approaches to modelling DE behaviour. A mathematical fractional Kelvin-Voigt model for DEA is presented in~\cite{Karner_2021_DEAPIDcontrol}, where they eliminate overshoot in PID control. In \cite{Cao_2018_DEAmodeling}, a spring-dashpots model describes the DEA's dynamics, with model parameters determined from the response of the robot.
FEM models for DE material have been explored, where deformation in unimoph DEA with inhomogeneous geometry modeled in~\cite{Araromi_2012_FEMDEA} uses piezoelectric elements. A simplified finite element analysis of a dielectric bending actuator is performed in~\cite{Zhou_2020_DEAFEM}. The FEM model for a gripping actuator in~\cite{Zhao_2008_FEMDE} is based on a custom defined material.
These methods focus on modelling the DEA behaviour in isolation (separate from the manipulation task). Thus, they lack an understanding of the task context in which the DEA manipulator is being used. In contrast, our method simulates the complete manipulation task, allowing us to learn both DEA behaviour and the task specific context.
}

An accurate simulator of the manipulation task can assist in learning a controller. For example, a position based dynamics simulator (defined using particle interactions) is used in~\cite{tataia_2017_pouring} to develop control strategies for pouring liquid. A FEM based differentiable simulator in~\cite{Prajjwal_2019_cutting} is used to learn control strategies for cutting.  The above methods are designed to control one specific manipulation task (\eg in~\cite{Heiden_2021_DiSECt}, the model is explicitly engineered for cutting). In contrast, our control framework can be used for \emph{any} robotic manipulation task that can be simulated in FEM.

\newtext{Learning based modelling and control can be used for soft robot manipulation. For example, a differentiable model of a soft robot’s quasi-static physics is learned in~\cite{Bern_2020_controlmodel}, and then used to perform gradient-based optimization to perform open-loop control. In \cite{George_2019_modelbasedcontrol}, a recurrent neural network is used as a fixed forward model within a policy learning algorithm for closed-loop soft robot control.} In contrast, our trained physics-informed model $f$ can infer properties of new \emph{unknown} setups, adapting during the closed-loop control process.

%%%%%%%%%%%%%%%%%%%%%%%%%%%%%%%%%%%%%%%%%%%%%%%%%%%%%%%%%%%%%%%%%%%%%%%%%%%%%%%%
%%%%%%%%%%%%%%%%%%%%%%%%%%%%%%%%%%%%%%%%%%%%%%%%%%%%%%%%%%%%%%%%%%%%%%%%%%%%%%%%
\begin{figure}[t]
    \centering
    \includegraphics[width=0.38\textwidth]{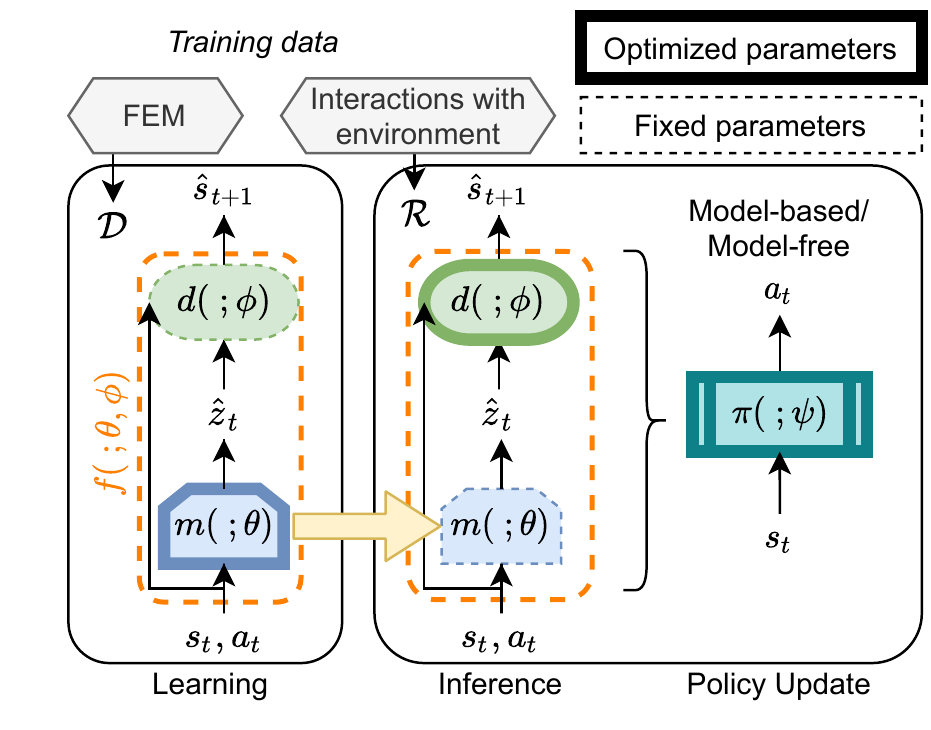}
    \caption{Training model $f$ - the simulator for our task. Material network $m$ is optimized during learning; dynamics $d$ inferred via closed-loop interaction.}
    \label{fig:trainingformodel}
\end{figure}
%%%%%%%%%%%%%%%%%%%%%%%%%%%%%%%%%%%%%%%%%%%%%%%%%%%%%%%%%%%%%%%%%%%%%%%%%%%%%%%%
%%%%%%%%%%%%%%%%%%%%%%%%%%%%%%%%%%%%%%%%%%%%%%%%%%%%%%%%%%%%%%%%%%%%%%%%%%%%%%%%
%%%%%%%%%%%%%%%%%%%%%%%%%%%%%%%%%%%%%%%%%%%%%%%%%%%%%%%%%%%%%%%%%%%%%%%%%%%%%%%%
\section{Control Framework}
\label{sec:framework}
%%%%%%%%%%%%%%%%%%%%%%%%%%%%%%%%%%%%%%%%%%%%%%%%%%%%%%%%%%%%%%%%%%%%%%%%%%%%%%%%
The objective of our control framework is to learn a control policy $\pi$ for a manipulation task.
\Fref{fig:trainingformodel} shows the closed-loop control design using the trained physics-informed model $f$ of the manipulation task along with policy $\pi$.

A model-based control approach utilizes a forward model of the system:
%$\mathcal{F}
$f:\mathcal{S}\times\mathcal{A}\rightarrow\mathcal{S}$, where $\mathcal{S}$ is the state space, and $\mathcal{A}$ is the control action space. For each timestep $t$,
the state is $s_t \in \mathcal{S}$, and the control action is $a_t \in \mathcal{A}$.
For MPC, the optimal control actions at each timestep is estimated by solving the optimization problem defined in~\eref{eq:MPC-opt-prob}, where $a_{\text{init}}$ is the initial action.
%The physics-informed model $f$ is
We particularly choose to use a physics-informed $f$, which is
differentiable, and allows us to use gradient-based methods to solve this optimization problem (such as finite-horizon Linear Quadratic Regulator, LQR~\cite{Brandon_2019_MPCDiff}). The objective of the control (\eg get to a target location) is used to define the cost function $\mathcal{C}:\mathcal{S}\times\mathcal{A}\rightarrow\mathbb{R}$ (\eg distance from the target location). For example, in DEA pulling, cost function $\mathcal{C}$ is defined for the objective of achieving target state (\ie the target location of the coin) with penalty on control actions to minimize actuation voltage of the DEA. \newtext{\eref{eqn:MPCcost} shows the cost function, where ${s}_t^\ast$ is target state, and $w_s, w_a$ the state and action penalties.}
%\footnote{The cost function defined in~\cite{Brandon_2019_MPCDiff}. For details, refer to~\sref{sec:parametersettging}.} 
\begin{equation}
    \begin{split}
        & \argmin_{s_{1:T}\in\mathcal{S},a_{1:T}\in\mathcal{A}}
        \sum_{t=1}^{T} \mathcal{C}(s_t,~a_t)\\
        &\text{s.t.} \quad %\ f: s_t,~a_t~\rightarrow~s_{t+1}~\&~a_1=a_{\text{init}} \\
        s_{t+1} = f(s_t, a_t) \quad \textrm{and} \quad a_1=a_\textrm{init}
    \end{split}
\label{eq:MPC-opt-prob}
\end{equation}
\begin{equation}
  \mathcal{C}(s_t,~a_t) = \frac{1}{2} %\sum\limits_{t=1}^{T} 
  (w_s(s_t - {s}_t^\ast)^2 + 
  %\frac{1}{2} %\sum\limits_{t=1}^{T} 
  w_aa_t^2)  \label{eqn:MPCcost}
\end{equation}

The physics of the robotic manipulation task includes, 
\begin{enumerate*}[label=(\roman*)]
    \item the physical laws of the deformable material behaviour, \eg electromechanical/hyperelastic behaviour, characterized by high order ODEs/PDEs that are computationally complex, and,
    \item the physical laws built according to the context of the manipulation task, \eg sliding motion laws, or gravity.
\end{enumerate*}
In modeling these physics, interaction variables $z$ are introduced to describe the contact properties (\eg force, stress, pressure) between the deformable material and its surroundings.
In particular, for DEA pulling, $z$ are the forces exerted by the DEA actuator on the contact surface. 

We simulate the manipulation task using a FEM model to numerically solve the associated physics equations. In this model, state $s_{t+1}$ and interaction variables $z_{t}$ are simulated, given $s_t$ and $a_t$.
The FEM model for DEA pulling is described in~\sref{sec:FEADEA}.
How the simulated data is then used to train a physics-informed model $f$ is described below.

%%%%%%%%%%%%%%%%%%%%%%%%%%%%%%%%%%%%%%%%%%%%%%%%%%%%%%%%%%%%%%%%%%%%%%%%%%%%%%%%
\subsection{Physics-informed model ($f$)}
\label{sec:phymodel}
%%%%%%%%%%%%%%%%%%%%%%%%%%%%%%%%%%%%%%%%%%%%%%%%%%%%%%%%%%%%%%%%%%%%%%%%%%%%%%%%
The physics-informed model $f$ has two parts:
\begin{enumerate}[(i)]
    \item The material network~($m$):
    A function approximator with weights {$\theta$} estimating interaction variables $\hat{z}$ (\eref{eqn:latentfunction}). These interaction variables ($z$) characterize deformable material behaviour in the manipulation task (\eg forces by DEA on contact surface).
    \item The dynamics~($d$): Physical laws characterizing the motion/dynamics of the system in the form of mathematical equations (\eg linear equations or ODEs/PDEs representing sliding or gripping). The dynamics $d$ estimate the next state using interaction variables $z$, state $s$, and action $a$~(\eref{eqn:systemdynamics}). Parameters $\phi$ describe the system's physical properties, \eg the mass of coin. 
\end{enumerate}
Thus, we can write the model $f$ as in~\eref{eqn:forwardmodel}. 
\begin{align} 
    \hat{s}_{t+1} &= f({s_t,~a_t;~\theta,~\phi}) \label{eqn:forwardmodel} \\
    f({s_t,~a_t;~\theta,~\phi}) &= d({\hat{z}_t,~s_t,~a_t;~\phi}) \label{eqn:systemdynamics}\\
    \hat{z}_{t} &= {m}({s_t,~a_t;~\theta})  \label{eqn:latentfunction}
\end{align}
\newtext{where $\theta$ are material neural network weights, and $\phi$ the parameters (\eg mass) for system dynamics.}
The material model $m$ of a deformable material can describe an actuator (\eg DEA), or the manipulated object (\eg cloth) depending on the manipulation task. \newtext{The interaction variables $z$ depend only on the deformable material (\ie the material network $m$), and are not impacted by the dynamics $d$ of the manipulation.}  For example, in DEA pulling, $m$ describes the actuator behaviour of a unimorph DEA (\sref{sec:FEADEA}). \newtext{Our model $f$ is a physics informed neural network~\cite{raissi_2019_pinn}, where physical rules can be imposed on interaction variables $z$ in the form of ODEs/PDEs. However, for the case study of DEA pulling, a system of linear equations sufficiently defines dynamics $d$.}

We next show how to define $f$ using physical laws. Model $f$ is differentiable, and captures the manipulation task physics through dynamics $d$. In addition to its use as a simulator to generate data, we also use it for inference, and for gradient-based control learning.

%%%%%%%%%%%%%%%%%%%%%%%%%%%%%%%%%%%%%%%%%%%%%%%%%%%%%%%%%%%%%%%%%%%%%%%%%%%%%%%%
\subsection{Training and policy synthesis}
%%%%%%%%%%%%%%%%%%%%%%%%%%%%%%%%%%%%%%%%%%%%%%%%%%%%%%%%%%%%%%%%%%%%%%%%%%%%%%%%
%%%%%%%%%%%%%%%%%%%%%%%%%%%%%%%%%%%%%%%%%%%%%%%%%%%%%%%%%%%%%
\small
\begin{algorithm}[t]
\caption{Learning}
\label{Algo:learning}
\begin{algorithmic}[1]
    \renewcommand{\algorithmicrequire}{\textbf{Input:}}
    \renewcommand{\algorithmicensure}{\textbf{Output:}}
    \ENSURE $\text{Material network }{m}$;\\
    \STATE $\text{Randomly initialize weights }\theta\text{, and fix parameters }\phi$;
    \WHILE {not stopping condition}
        \STATE $a_t \leftarrow \text{select action at $t$ using a fixed policy}$;\\
        \STATE $s_{t+1},z_{t} = \mathit{FEM}(s_{t},~a_{t})$;\\
        \STATE $\text{Dataset }\mathcal{D} \leftarrow \mathcal{D} \cup (s_t, a_t, z_t, s_{t+1})$;
    \ENDWHILE
    %\WHILE{not stopping condition}
        \STATE Using all data from $\sim{\mathcal{D}}$; {// Say $|\mathcal{D}| = N$ data samples} \\
        \STATE $\hat{z}_{t} = {m}(s_{t},~a_{t};~\theta)$;\\
        \STATE $\hat{s}_{t+1} = f(s_{t},~a_{t};~\theta,~\phi)$;\\
        \STATE $\theta \leftarrow \theta - \alpha_{\theta} \frac{d \mathcal{L}_{l}(\theta,~\phi)}{d\theta}$; 
        {// Update $\theta$ using $\hat{z}$ and $\hat{s}$,~\eref{eqn:loss_learning}}
    %\ENDWHILE
    \RETURN ${m}$
\end{algorithmic} 
\end{algorithm}
\normalsize
%%%%%%%%%%%%%%%%%%%%%%%%%%%%%%%%%%%%%%%%%%%%%%%%%%%%%%%%%%%
%%%%%%%%%%%%%%%%%%%%%%%%%%%%%%%%%%%%%%%%%%%%%%%%%%%%%%%%%%%%%
\small
\begin{algorithm}[t!]
 \caption{Control}
 \label{Algo:control}
 \begin{algorithmic}[1]
 \renewcommand{\algorithmicrequire}{\textbf{Input:}}
 \renewcommand{\algorithmicensure}{\textbf{Output:}}
 \REQUIRE $\text{Trained weights }\theta$
 \STATE $\text{Randomly initialize parameters }\phi \text{, and } \psi \text{, and fix }\theta$;
 \STATE $s_{1} \leftarrow \mathit{env.reset}()$;\\
 \WHILE {not stoping condition}
    \STATE $a_{t} = \pi(s_t;~\psi)$;\\
    \STATE $s_{t+1}, r_{t} \leftarrow \mathit{env.step}(a_{t})$;\\
    \STATE $\text{Replay buffer }\mathcal{R} \leftarrow \mathcal{R} \cup (s_t, a_t, r_t, s_{t+1})$;
    \IF {it's time to update}
     \STATE $\text{Randomly sample } B\text{ transitions from} \sim{\mathcal{R}}$;\\
     \STATE \textit{// Inference}\\
     \STATE $\hat{s}_{t+1} = f(s_{t},~a_{t};~\theta,~\phi)$;\\
     \STATE $\phi \leftarrow \phi - \alpha_{\phi} \frac{d \mathcal{L}_{i}(\theta,~\phi)}{d\theta}$; {{// Update $\phi$ using $\hat{s}$,~\eref{eqn:loss_inference}}}\\
     \STATE \textit{// Policy Update}\\
     \STATE $\text{Update $\psi$ by policy defined updates,~\eg SAC}$~\cite{haarnoja_2018_SAC};\\
     \ENDIF
 \ENDWHILE 
 \end{algorithmic} 
 \end{algorithm}
 \normalsize
%%%%%%%%%%%%%%%%%%%%%%%%%%%%%%%%%%%%%%%%%%%%%%%%%%%%%%%%%%%

The physics-informed model $f$ and policy $\pi$ are learnt in two steps.
First, a \emph{Learning} step optimizes weights $\theta$ of ${m}$, using data generated by the FEM model of the task.
Second, a \emph{Control} step, where policy $\pi$ is learnt via environment interactions. 
These interactions are used to infer parameters $\phi$  (\eg coin mass) of dynamics $d$, informing $f$. We then use $f$ to learn $\pi$.
%%%%%%%%%%%%%%%%%%%%%%%%%%%%%%%%%%%%%%%%%%%%%%%%%%%%%%%%%%%%%%%%%%%%%%%%%%%%%%%%
\paragraph{Learning}
%%%%%%%%%%%%%%%%%%%%%%%%%%%%%%%%%%%%%%%%%%%%%%%%%%%%%%%%%%%%%%%%%%%%%%%%%%%%%%%%
The weights $\theta$ are optimized by minimizing the loss function based on the error in estimating material model, \ie $m$: ($z_{t} - \hat{z}_{t}$), and the error in enforcing dynamics, \ie $d$: ($s_{t} - \hat{s}_{t}$). Incorporating the loss encountered in $\hat{s}$ ensures that our model adheres to dynamics $d$ (as shown by optimization of physics informed neural networks~\cite{raissi_2019_pinn}).
\algoref{Algo:learning} shows how $\theta$ is optimized by fixing parameters $\phi$ and minimizing the learning loss $\mathcal{L}_l$ (\eref{eqn:loss_learning}).
A fixed policy is used to select actions $a_t$ (\eg a random or uniform policy). The learning rate for weights $\theta$ is $\alpha_{\theta}$ and number of data samples is $N$.

\begin{equation}
  \mathcal{L}_l(\theta,\phi) = \frac{1}{N} \sum\limits_{t=1}^{N} 
  (z_t - \hat{z}_t)^2 + 
  \frac{1}{N} \sum\limits_{t=1}^{N} 
  (s_t - \hat{s}_t)^2 \label{eqn:loss_learning}
\end{equation}

%%%%%%%%%%%%%%%%%%%%%%%%%%%%%%%%%%%%%%%%%%%%%%%%%%%%%%%%%%%%%%%%%%%%%%%%%%%%%%%%
\paragraph{Control}
\label{sec:frameworkcontroller}
%%%%%%%%%%%%%%%%%%%%%%%%%%%%%%%%%%%%%%%%%%%%%%%%%%%%%%%%%%%%%%%%%%%%%%%%%%%%%%%%
\algoref{Algo:control} shows our steps for closed-loop control. First, in inference, parameters $\phi$ are optimized by minimizing the dynamics estimation error, \ie $d$: ($s_{t} - \hat{s}_{t}$). \eref{eqn:loss_inference} gives the loss function, with learning rate $\gamma_{\phi}$ and batch size $B$.
\begin{equation}
  \mathcal{L}_i(\theta,\phi) = \frac{1}{B} %T_c}
  \sum\limits_{t=1}^B %{T_c} 
  (s_t - \hat{s}_t)^2  \label{eqn:loss_inference}
\end{equation}

Second, we learn policy $\pi$ with weights $\psi$. Updates in $\psi$ are defined using the underlying policy $\pi$ and its training objective. For the model-free policy trained using the SAC algorithm, $\psi$ is updated based on the loss function from~\cite{haarnoja_2018_SAC}; the weights $\psi$ represent weights of the neural network.
For model-based MPC, we use our trained model $f$ and gradient-based optimization to estimate the best action. 
Thus, with MPC, updating parameters~$\psi$ (line~13) is unnecessary.

%\input{Sections/Appendix 2}
%%%%%%%%%%%%%%%%%%%%%%%%%%%%%%%%%%%%%%%%%%%%%%%%%%%%%%%%%%%%%%%%%%%%%%%%%%%%%%%%
%%%%%%%%%%%%%%%%%%%%%%%%%%%%%%%%%%%%%%%%%%%%%%%%%%%%%%%%%%%%%%%%%%%%%%%%%%%%%%%%
\section{Soft Robotic DEA pulling}
\label{sec:roboticpullingsetup}
%%%%%%%%%%%%%%%%%%%%%%%%%%%%%%%%%%%%%%%%%%%%%%%%%%%%%%%%%%%%%%%%%%%%%%%%%%%%%%%%
%%%%%%%%%%%%%%%%%%%%%%%%%%%%%%%%%%%%%%%%%%%%%%%%%%%%%%%%%%%%%%%%%%%%%%%%%%%%%%%%
%%%%%%%%%%%%%%%%%%%%%%%%%%%%%%%%%%%%%%%%%%%%%%%%%%%%%%%%%%%%%%%%%%%%%%%%%%%%%%%%
\begin{figure}[t!]
    \centering
    \includegraphics[width=0.4\textwidth]{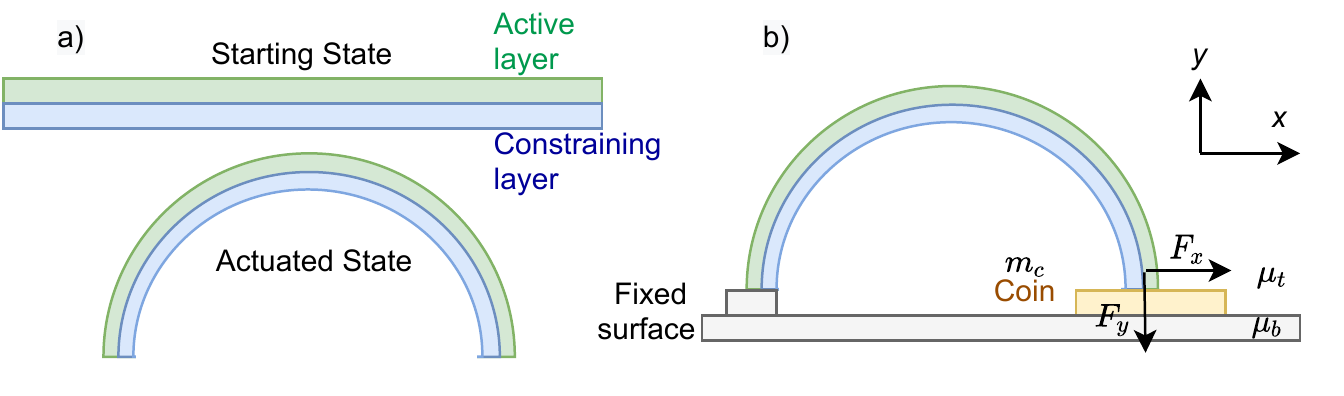}
    \caption{(a)~Unimorph DEA: Active layer expands under influence of electric voltage, causing bending. (b)~Problem setup: DEA moves stationary coin.}
    \label{fig:DEADiagrams}
\end{figure}
%%%%%%%%%%%%%%%%%%%%%%%%%%%%%%%%%%%%%%%%%%%%%%%%%%%%%%%%%%%%%%%%%%%%%%%%%%%%%%%%
%%%%%%%%%%%%%%%%%%%%%%%%%%%%%%%%%%%%%%%%%%%%%%%%%%%%%%%%%%%%%%%%%%%%%%%%%%%%%%%%
% \subsection{Pulling Setup ($S$)}
% \label{sec:setup}
%%%%%%%%%%%%%%%%%%%%%%%%%%%%%%%%%%%%%%%%%%%%%%%%%%%%%%%%%%%%%%%%%%%%%%%%%%%%%%%%
We design the manipulation task of coin pulling using a unimorph Dielectric Elastomers Actuators (DEAs) to evaluate the framework proposed in \sref{sec:framework}. The deformable DEA actuator is made of Dielectric Elastomers (DEs), which are a type of electroactive polymers that produce mechanical strain under the influence of electric voltage. Thus, a DE membrane expands its area when a voltage is applied across its thickness~\cite{Pelrine_1998_DEA}. 
%A typical DEA consists of polymer film that is coated with compliant electrodes on both sides. Multiple layers of active and inactive DE membranes can be stacked to design unimorph or bimorph DEA.

\Fref{fig:DEADiagrams}(a) shows a unimorph DEA, with one active and one constraining layer. The active layer expands under externally applied voltage causing the bending motion.
The DEA is fixed at one end, and the other end rests freely on a circular coin $c$. On actuation, the DEA acts as a soft robotic finger, pulling the coin. A controller policy $\pi$ can be learnt to achieve a certain displacement in the coin.
\Fref{fig:DEADiagrams}(b) shows the 2D view of the setup, where the mass of the coin is $m_c$, the kinetic friction coefficient between the coin and DEA is $\mu_t$, and the kinetic friction coefficient between the coin and bottom surface is $\mu_b$.
The displacement of the coin depends on such parameters of the system. A pulling coin setup $C_c$ is characterized by fixed values of $\{m_c, \mu_t, \mu_b\}$. Setups $C_1, C_2, \dots$ represent pulling different coins, based on different parameter values.\footnote{We consider the coins to be of fixed dimensions (\ie fixed volume), and thus change the mass $m_c$ by changing the density $\rho$ of the coin material. For further details, please refer to~\sref{sec:simresults}.}

The physics-informed model $f$ of the system is defined by the variables shown in~\tref{tab:parametdefinations}. 
The state of the system at time-step $t$ is characterized by the location $x_t$ and velocity $u_t$ of the coin along the $x$-axis.
The action comprises the voltage ($V_t$) applied on the DEA and $\Delta t$,\footnote{Note that in FEM, the time between successive simulation datapoints may vary. \newtext{Thus, $\Delta t$ is variable during training the model $f$.}} and the hidden variables are the forces ($F_x$ and $F_y$) applied by the DEA on the top surface of the coin.

%%%%%%%%%%%%%%%%%%%%%%%%%%%%%%%%%%%%%%%%%%%%%%%%%%%%%%%%
\begin{table}[t]
% \centering
% \ra{1.2}
% \caption{State and action definition for timestep t}
% \begin{tabular}{@{\extracolsep{\fill}}ccl@{}}
% \multirow{2}{*}{s$_{t}$} & x$_{t}$ & Location of the coin along $x$-axis at time $t$ \\
% & u$_{t}$ &  Velocity of the coin along $x$-axis at time $t$  \\ [3pt] 
% \multirow{2}{*}{a$_{t}$}  & V$_{t}$ & Voltage applied on the DEA\\
% & $\Delta{t}$ & time difference between t and t+1 \\ [3pt] 
% \multirow{2}{*}{z$_{t}$} & F$_{x,t}$ & Force along $x$-axis by DEA on coin $c$ \\
% & F$_{y,t}$ & Force along $y$-axis by DEA on coin $c$ \\[3pt] 
% \end{tabular}
% \label{tab:parametdefinations}
\centering
\ra{1.2}
\caption{State and action definition for timestep \emph{t}}
\begin{tabular}{@{\extracolsep{\fill}}c|cl@{}}
\hline
\multirow{2}{*}{s$_{t}$} & x$_{t}$ & Location of coin along $x$-axis at time $t$ \\
& u$_{t}$ &  Velocity of coin along $x$-axis at time $t$  \\  [3pt] 
\multirow{2}{*}{a$_{t}$}  & V$_{t}$ & Voltage applied on the DEA\\
& $\Delta{t}$ & time difference between t and t+1 \\ [3pt] 
\multirow{2}{*}{z$_{t}$} & F$_{x,t}$ & Force along $x$-axis by DEA on coin $c$ \\
& F$_{y,t}$ & Force along $y$-axis by DEA on coin $c$ \\[3pt] 
\hline
\end{tabular}
\label{tab:parametdefinations}
\end{table}
%%%%%%%%%%%%%%%%%%%%%%%%%%%%%%%%%%%%%%%%%%%%%%%%%%%%%%%%

%%%%%%%%%%%%%%%%%%%%%%%%%%%%%%%%%%%%%%%%%%%%%%%%%%%%%%%%%%%%%%%%%%%%%%%%%%%%%%%%
\subsection{Material network ($m$)}
%%%%%%%%%%%%%%%%%%%%%%%%%%%%%%%%%%%%%%%%%%%%%%%%%%%%%%%%%%%%%%%%%%%%%%%%%%%%%%%%
Modeling non-linear properties of DEs require modeling the effects of hyperelasticity and Maxwell stress~\cite{Pelrine_1998_DEA}. On application of voltage $V,$ maxwell stress causes the bending actuation in DEA. The actuated DEA exerts forces $F_x$ and $F_y$ on the top surface of the coin, which results in its motion. The material network used to estimate these forces is defined in \eref{eqn:latentDEA}. We simulate DEA pulling using FEM, to generate data and optimize weights $\theta$ (\sref{sec:FEADEA}).
\begin{align} 
    \hat{F}_{x,t}, \hat{F}_{y,t} &= m({{x}_{t},~{u}_{t},~{V}_{t},~{\Delta t};~\theta})  \label{eqn:latentDEA}
\end{align}

%%%%%%%%%%%%%%%%%%%%%%%%%%%%%%%%%%%%%%%%%%%%%%%%%%%%%%%%%%%%%%%%%%%%%%%%%%%%%%%%
\subsection{Dynamics ($d$)}
\label{sec:SetupD}
%%%%%%%%%%%%%%%%%%%%%%%%%%%%%%%%%%%%%%%%%%%%%%%%%%%%%%%%%%%%%%%%%%%%%%%%%%%%%%%%
Physical laws of the pulling setup define the system dynamics $d$ (\sref{sec:phymodel}). There are two stages during pulling: \emph{static friction} (forces are applied but there is no motion), and \emph{kinetic friction} (applied forces cause motion in coin). An actuation threshold voltage $V^{T}$ is required to achieve a minimum coin displacement (\ie to get to the stage of kinetic friction). Coin acceleration $A_t$ is due to the net force in the $x$-axis (\eref{eqn:netforce}), where F$_{\mu}$ is the opposing frictional force. We calculate F$_{\mu}$ using~\eref{eqn:Fmu}, assuming linear growth in frictional force during the stage of static friction, and a no-slip condition on the top surface. 
\begin{align}
\hat{F}_{x,t}^{Net} & = m_{c} A_{t} = \hat{F}_{x,t} - {F}_{\mu,t}
\label{eqn:netforce}
\end{align}
\begin{equation}
  F_{\mu,t} =
\begin{dcases*}
\mu_{b} \left(\hat{F}_{y,t} \Plus m_c g\right) & $\text{if}~{V \geq V^{T}}$;\\
\frac{\mu_{b}V_t}{V^T} \left(\hat{F}_{y,t} \Plus m_c g\right) & $\text{otherwise}$
\end{dcases*}
\label{eqn:Fmu}
\end{equation}

where \textit{g} is gravitational acceleration (9.8\,m/s$^2$). We assume a frictional velocity decay for a moving coin if DEA actuation is stopped (\ie $V_t$ = 0). These dynamics disregard non-linear motion in coin (with high DEA actuation voltages, where DEA loses contact with the coin surface). 
The dynamics $d$ of this pulling setup is characterized by parameters $\phi$ (\sref{sec:framework}), which are: coin mass $m_c$ and frictional coefficient $\mu_b$. While training $m$, these values are fixed. We infer $\phi$ during closed-loop environment interactions.

The next location and velocity of the coin, \ie $\hat{x}_{t+1}$ and $\hat{u}_{t+1}$, are calculated using $A_t$, $x_t$, $u_t$, and $\Delta t$ and equations of motion. Thus, the dynamics $d$ is a set of linear equations based on the laws of motion.

%%%%%%%%%%%%%%%%%%%%%%%%%%%%%%%%%%%%%%%%%%%%%%%%%%%%%%%%%%%%%%%%%%%%%%%%%%%%%%%%
%%%%%%%%%%%%%%%%%%%%%%%%%%%%%%%%%%%%%%%%%%%%%%%%%%%%%%%%%%%%%%%%%%%%%%%%%%%%%%%%
\section{FEM of DEA pulling}
\label{sec:FEADEA}
%%%%%%%%%%%%%%%%%%%%%%%%%%%%%%%%%%%%%%%%%%%%%%%%%%%%%%%%%%%%%%%%%%%%%%%%%%%%%%%%
%%%%%%%%%%%%%%%%%%%%%%%%%%%%%%%%%%%%%%%%%%%%%%%%%%%%%%%%%%%%%%%%%%%%%%%%%%%%%%%%
%%%%%%%%%%%%%%%%%%%%%%%%%%%%%%%%%%%%%%%%%%%%%%%%%%%%%%%%%%%%%%%%%%%%%%%%%%%%%%%%
\begin{figure}[t!]
    \centering
    \includegraphics[width=0.35\textwidth]{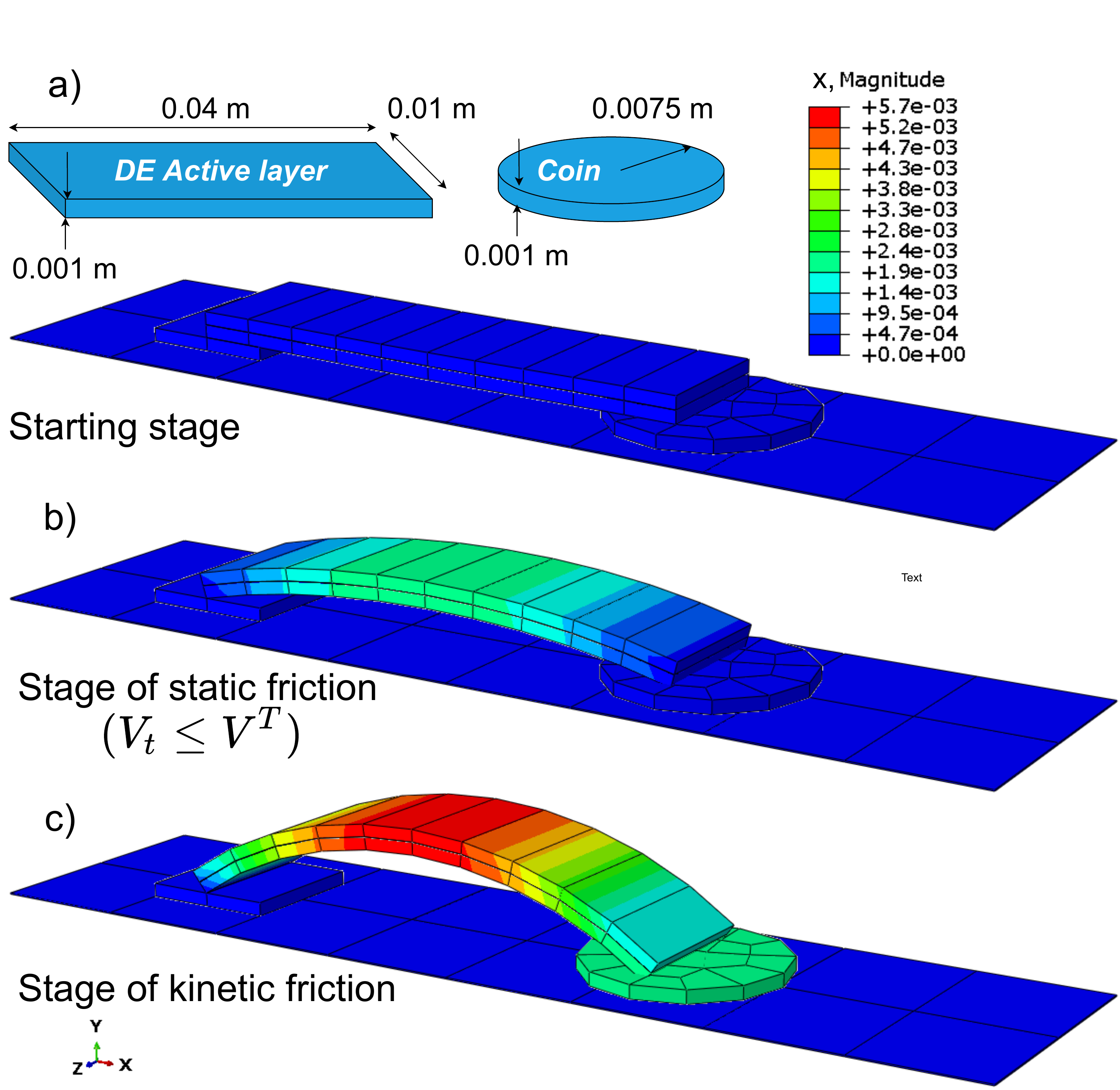}
    \caption{FEM of DEA pulling, (a)~inactive DEA with dimensions, (b)~active DEA during static friction stage (c)~motion occurs in kinetic friction stage.}
    \label{fig:FEMimages}
\end{figure} 
%%%%%%%%%%%%%%%%%%%%%%%%%%%%%%%%%%%%%%%%%%%%%%%%%%%%%%%%%%%%%%%%%%%%%%%%%%%%%%%%

FEM is a numerical method for solving differential equations of a physical system. \newtext{Its complexity depends on both the task and available computational power.} The physical system of DEA pulling consists of a unimorph DEA on the fixed surface and a solid coin (\sref{sec:roboticpullingsetup}). We use commercially available software ABAQUS~\cite{Smith_2009_abaqus} to build a 3D model of DEA pulling.
\Fref{fig:FEMimages} shows the simulated setup, during the starting stage and the stages of static and kinetic friction. In the stage of static friction, we clearly see no motion in the coin even when actuating the DEA.

%%%%%%%%%%%%%%%%%%%%%%%%%%%%%%%%%%%%%%%%%%%%%%%%%%%%%%%%%%%%%%%%%%%%%%%%%%%%%%%%
\paragraph{DE Material}
\label{sec:DEMaterialFEM}
%%%%%%%%%%%%%%%%%%%%%%%%%%%%%%%%%%%%%%%%%%%%%%%%%%%%%%%%%%%%%%%%%%%%%%%%%%%%%%%%
\newtext{To model the non-linear elastic behaviour in DE membranes, recent works use hyperelastic material models, like the Gent and the Neo-Hookean models~\cite{Zhang_2019_DEAnonlinear}. For the small strains present in the thin membrane of our DEA ($\leq10\%$), such hyperelastic  behaviour reduces to linear elasticity.} Furthermore, since no commercial FEM packages provide DE elements out of the box, we approximate the behaviour of our DE material in FEM using piezoelectric materials elements~\cite{Piefort_2000_FEMPeiso}.
We modify the piezoelectric finite elements material properties to model the Maxwell stress effect observed in dielectric materials. The Maxwell stress $p$ on the DE membrane is given by \eref{eqn:DEAstrain}~\cite{Pelrine_1998_DEA}.
Similar to DE, piezoelectric materials exhibit strain when in the presence of electric fields. The piezoelectric stress is given by~\eref{eqn:peizoelectricstress}.\begin{equation}
    \label{eqn:DEAstrain}
    p = {e_0e_r}\Big(\frac{V}{z}\Big)^2
\end{equation}
\begin{equation}
\label{eqn:peizoelectricstress}
\sigma_{ij} = D^E_{ijkl}\epsilon_{kl} - e_{mij}E_m
\end{equation}

where $V$ is the applied voltage across thickness $z$ of the DE membrane, the relative permittivity is $e_r$, and the permittivity of free space is $e_0$. The piezoelectric elastic stiffness matrix is $D^E_{ijkl}$, the strain tensor is $\epsilon_{kl}$, the stress coefficient is $e_{mij}$ and the electric potential gradient is $E_m$. As detailed in~\cite{Araromi_2012_FEMDEA} the piezoelectric stress becomes approximately equal to the Maxwell stress for a thin membrane, such that the strain $e_{zz}$ in the direction of thickness ($z$-axis) is given by:
\begin{equation}
\label{eqn:peizoelectricstrain}
e_{zz} = e_re_0E_z
\end{equation}

%\oldtext{These piezoelectric elements assume linear elasticity, which is not a limiting factor as we can assume such behaviour for our case of thin DE membranes~\cite{Araromi_2012_FEMDEA}.}

%%%%%%%%%%%%%%%%%%%%%%%%%%%%%%%%%%%%%%%%%%%%%%%%%%%%%%%%%%%%%%%%%%%%%%%%%%%%%%%%
\subsection{FEM simulation settings}
\label{sec:FEMsimulationsettings}
%%%%%%%%%%%%%%%%%%%%%%%%%%%%%%%%%%%%%%%%%%%%%%%%%%%%%%%%%%%%%%%%%%%%%%%%%%%%%%%%

%Piezoelectric material elements are used to define the material for the unimorph DEA.
\Fref{fig:FEMimages}(a) shows our FEM setup when the DEA is inactive (zero voltage). The mesh consists of an 8-node linear brick (ABAQUS element type C3D8E). Each DE membrane has 10 elements.\footnote{We limit the number of mesh elements due to software limitations.} For meshing the coin, ABAQUS's internal meshing strategy is used to generate 20 elements. 

For both the active and constraining layers of the DE material, the Poisson’s ratio is 0.5 and Young’s modulus is 0.56 MPa~\cite{Wang_2017_DEProp}. We use~\eref{eqn:peizoelectricstrain} to calculate $e_{zz}$ = 3.68. All other piezoelectric coefficients are zero. We assume elastic behaviour for the coin. The bottom surface and the fixed end of the DEA are constrained using encastre boundary condition. The top surface assumes a no-slip condition. 

The coin rests on the frictional surface with coefficient $\mu_b \in$ \{0.2, 0.25\} (defined as tangential behaviour in the contact interactions in ABAQUS). Similarly, the free end of the DEA rests atop the coin with $\mu_t \in$ \{0.5, 0.55\}. To simulate a real scenario, we include gravitational load ($g$ = 9.8\,m/s$^\textrm{2}$). We also do not assume a no-slip condition between the top of the coin and DEA, in contrast to the dynamics in~\sref{sec:SetupD}. Coin mass ($m_c$) is calculated using its volume and density $\rho \in$ \{7.7,~7.8\}\,g/cm$^\textrm{3}$, thus, making $m_c \in$ \{1.36,~1.38\}\,g.
%1.37837
%1.36070
The total time simulated in FEM is 1\,s, with $\Delta t$ between points determined by the internal solver.

%%%%%%%%%%%%%%%%%%%%%%%%%%%%%%%%%%%%%%%%%%%%%%%%%%%%%%%%%%%%%%%%%%%%%%%%%%%%%%%%
%%%%%%%%%%%%%%%%%%%%%%%%%%%%%%%%%%%%%%%%%%%%%%%%%%%%%%%%%%%%%%%%%%%%%%%%%%%%%%%%
%\section{Experimental Design}
%\label{sec:exptdesign}
%%%%%%%%%%%%%%%%%%%%%%%%%%%%%%%%%%%%%%%%%%%%%%%%%%%%%%%%%%%%%%%%%%%%%%%%%%%%%%%%
%%%%%%%%%%%%%%%%%%%%%%%%%%%%%%%%%%%%%%%%%%%%%%%%%%%%%%%%%%%%%%%%%%%%%%%%%%%%%%%%

%%%%%%%%%%%%%%%%%%%%%%%%%%%%%%%%%%%%%%%%%%%%%%%%%%%%%%%%%%%%%%%%%%%%%%%%%%%%%%%%
\section{Parameters settings}
\label{sec:parametersettging}
%%%%%%%%%%%%%%%%%%%%%%%%%%%%%%%%%%%%%%%%%%%%%%%%%%%%%%%%%%%%%%%%%%%%%%%%%%%%%%%%
\newtext{This section details the architectures, parameters, and training procedures used to evaluate our framework.}
An experimental setup $C$ is defined using differing values of $\langle \mu_t, \mu_b, m_c \rangle$ (\sref{sec:roboticpullingsetup}). We use 8 setups, denoted $\{C_1, C_2, \dots, C_8\}$. An FEM model is developed for each setup. To collect the dataset for each model, we apply a linearly increasing electric potential load ($V_t \in \{0.0, 400.0\}V$) to the top surface of the active DE layer. \tref{tab:parametdefinations} describes the values collected at each timestep~$t$. Each dataset contains 1000-2000 data points. Across setups, $\langle x_0, u_0 \rangle = \langle 0.0, 0.0 \rangle$. 

During learning, we initialize parameters $\phi$ ($m_c$ and $\mu_b$, \sref{sec:SetupD}) of dynamics $d$ using the true values from the FEM model. 
Note that $\mu_t$ is \emph{not} used as a parameter in the dynamics of 
$f$, but, is required for the FEM modeling. \newtext{For each setup, the threshold voltage ($V^T$) is the voltage required to achieve a displacement of 
$-$10$^{-\textrm{5}}$\,m.} We assume the coin loses contact with the DEA for $V_t$ $\ge$ 300\,V. 

Physics-informed model $f$ is developed using Pytorch~\cite{paszke_2017_pytorch} and contains the material network $m$ and the dynamics $d$ (\sref{sec:framework}). The material network $m$ is a fully connected neural network with four input nodes, two output nodes, three hidden layers with \newtext{128} neurons each, and rectified linear (ReLU) activation functions. An ADAM optimizer with learning rate $0.001$ is used to minimize \eref{eqn:loss_learning} and \eref{eqn:loss_inference} for 1,000 iterations. An early stopping criterion is used based on validation loss with 0.0 minimum change. \newtext{The baseline neural network uses RelU activations, while the recurrent neural network uses $L\!S\!T\!M$ cells. The $L\!S\!T\!M$ uses an 8-step recurrence.}

The target state for the controller is {$x^T$} = $-$1.0\,mm, \ie goal is to achieve a 1\,mm displacement. 
\newtext{The manipulation `episode' is simulated, and terminates when $\|x_t-x^T\| <$ 0.01\,mm. 
To simulate real-world discrepancies, we add gaussian noise to simulated state $x_t$ with zero mean and standard deviation of 0.001\,m.
}
We average results for 10 `episodes' for all controllers. Batch size (\algoref{Algo:control}) is 256. \newtext{During control, the time increment $\Delta t$ is fixed to 0.001\,s---the policy does not provide it as an output}. For inference, $m_c$ is initialized to 0.001\,g and $\mu_b$ is initialized to 0.2.

\newtext{We use the model-based control policy from~\cite{Brandon_2019_MPCDiff}}. For MPC, we set the number of timesteps 20, LQR iterations to 20, and action penalty to 0.001. The model-free policy is trained using Soft Actor-Critic (SAC)~\cite{haarnoja_2018_SAC}. For SAC, fully connected neural networks are used for actor and critic with two hidden layers of 256 neurons each. The value of $\tau$ (soft updates) is set to 0.005, and networks are optimized using MSE loss and the ADAM optimizer. For the PD controller, the value of $K_p$ is set to $-$0.5 and $K_d$ is set to 5.

%%%%%%%%%%%%%%%%%%%%%%%%%%%%%%%%%%%%%%%%%%%%%%%%%%%%%%%%%%%%%%%%%%%%%%%%%%%%%%%%
%%%%%%%%%%%%%%%%%%%%%%%%%%%%%%%%%%%%%%%%%%%%%%%%%%%%%%%%%%%%%%%%%%%%%%%%%%%%%%%%
\section{Experimental Results}
\label{sec:experimentalresults}
%%%%%%%%%%%%%%%%%%%%%%%%%%%%%%%%%%%%%%%%%%%%%%%%%%%%%%%%%%%%%%%%%%%%%%%%%%%%%%%%
%%%%%%%%%%%%%%%%%%%%%%%%%%%%%%%%%%%%%%%%%%%%%%%%%%%%%%%%%%%%%%%%%%%%%%%%%%%%%%%%

This section presents experiments describing the high simulation and inference accuracy of $f$, while also developing an effective closed-loop soft robotic controller. 
For the case of DEA pulling, we evaluate,
\begin{enumerate*}[label=(\roman*)]
    \item the accuracy of the $f$ as a simulator, and
    \item the accuracy of $f$ in inference, and
    \item the closed-loop MPC controller that utilizes $f$
\end{enumerate*}.
%We show that the physics-informed model learnt in our framework is an accurate simulator and can assist in fast closed-loop model-based control. 

%%%%%%%%%%%%%%%%%%%%%%%%%%%%%%%%%%%%%%%%%%%%%%%%%%%%%%%%%%%%%%%%%%%%%%%%%%%%%%%%
\subsection{Simulation (\qref{q:simulation})}
\label{sec:simresults}
%%%%%%%%%%%%%%%%%%%%%%%%%%%%%%%%%%%%%%%%%%%%%%%%%%%%%%%%%%%%%%%%%%%%%%%%%%%%%%%%

%%%%%%%%%%%%%%%%%%%%%%%%%%%%%%%%%%%%%%%%%%%%%%%%%%%%%%%%
\iffalse
\begin{table}[t]
\centering
\caption{6 Simulation Experiment Sets}
\begin{tabular}{@{\extracolsep{\fill}}c|c c c c c c}
        \hline
      {} & {1} & {2} & {3} & {4} & {5} & {6} \\
      \hline
      $C_{\text{train}}$ & $C_1$ & $C_2$ & $C_3$ & $C_4$ & $C_5$ & $C_6$\\
      $C_{\text{val}}$ & $C_2$ & $C_3$ & $C_4$ & $C_5$ & $C_6$ & $C_7$\\
      $C_{\text{test}}$ & $C_3$ & $C_4$ & $C_5$ & $C_6$ & $C_7$ & $C_8$ \\
      \hline
    \end{tabular}
    \label{tab:Exp1allsets}
\end{table}
\fi
%%%%%%%%%%%%%%%%%%%%%%%%%%%%%%%%%%%%%%%%%%%%%%%%%%%%%%%%

%%%%%%%%%%%%%%%%%%%%%%%%%%%%%%%%%%%%%%%%%%%%%%%%%%%%%
%\paragraph{\begin{experiment}\label{exp:simulation} Simulation\end{experiment}}
%%%%%%%%%%%%%%%%%%%%%%%%%%%%%%%%%%%%%%%%%%%%%%%%%%%%%

\newtext{We design an experiment to answer~\qref{q:simulation}, \ie} 
to show that $f$ can simulate data for new parameter settings, we train and simulate on different setups (defined in \sref{sec:roboticpullingsetup}). Note that different setups represent different coins, with different mass $m_c$ and frictional coefficients $\mu_b$ and $\mu_t$.

A simulation experiment set has coin setups given by $\{C_{\text{train}}, C_{\text{val}}, C_{\text{test}}\}$. FEM data from \{$C_{\text{train}}$, $C_{\text{val}}$\} is used to optimize weights $\theta$ (material network).  \newtext{The objective of learning step (\algoref{Algo:learning}) is to optimize weights $\theta$ of the material network $m$. During training and validation, the material network $m$ is same across coin setups (\ie same $\theta$ for $C_{\text{train}}$ and $C_{\text{val}}$) and dynamics $d$ are specific to coin setups (\ie $\phi$ based on $C_{\text{train}}$ and $C_{\text{val}}$).}
During testing, data is simulated recursively (for $T$ = 1\,s) in a test setup $C_{\text{test}}$, using simulator $f$. This $f$ consists of
\begin{enumerate*}[label=(\roman*)]
\item a material network with previously optimized $\theta$, and
\item the dynamics with parameters $\phi_{test}$ (based on $C_{\text{test}}$)
\end{enumerate*}. 
We present averaged results for 6 experiment sets %\oldtext{presented in \tref{tab:Exp1allsets}} 
\newtext{(E.g., for first set, $\{C_{\text{train}}, C_{\text{val}}, C_{\text{test}}\}$ are $\{C_1,~C_2,~C_3\}$, the second are $\{C_2,~C_3,~C_4\}$, etc.).}
%A simulation experiment uses three setups, given by $\{C_{train}, C_{val}, C_{test}\}$ representing the training setup, the validation setup, and the test setup. FEM data generated from $C_{train}$ and $C_{val}$ is used as training and validation data (to optimize $\theta$ of the material network). A total of 6 simulation experiments are defined using the 8 setups by consecutively choosing the next setup. For example, first simulation experiment has $\{C_1,~C_2,~C_3\}$, second simulation experiment has $\{C_2,~C_3,~C_4\}$, \etc
%The model $f_{test}$ has optimized $\theta$ and values of $\phi$ based on $C_{test}$. This $f_{test}$ is used to simulate data for $C_{test}$ for simulation 
%time $T$ (=1 s) recursively, where we use the real action~${a_t}$ and simulated state~${\hat{s}_t}$ at timestep $t$, to simulate the state ${\hat{s}_{t+1}}$ at next timestep~$t+1$. 

We evaluate the absolute errors encountered in simulating $\hat{s}_{t+1}$ ($x_t$ and $u_t$) at each timestep $t$. For example, error in $x_t$ is given by $e_t^{x} = |\hat{x}_{t} - {x}_{t}|$, where $\hat{x}_{t}$ is the location simulated using $f$, and ${x}_{t}$ is the true value (from the FEM dataset). Absolute errors in data simulated using a black-box baseline Neural Network ($N\!N$) \newtext{and long short term memory recurrent neural network ($L\!S\!T\!M$)} trained using data from $C_{\text{train}}$ and $C_{\text{val}}$ are also included (\eg a $N\!N$ that simply approximates $\hat{s}_{t+1} = N\!N(s_t,~a_t; w)$).

%%%%%%%%%%%%%%%%%%%%%%%%%%%%%%%%%%%%%%%%%%%%%%%%%%%%%%%%%%%%%%%%%%%%%%%%%%%%%%%%
% \begin{figure}[t!]
%     \centering
%     \includegraphics[width=0.4\textwidth]{Figures/Simulated-Location-and-Forces-1.pdf}
%     \caption{Data for setup $C_3$ ($\rho = 7.7\,\textrm{g}/\textrm{cm}^3, \mu_b = 0.2, \mu_t = 0.55$). Data is simulated given the action sequences in the test setup.}
%     \label{fig:openloopsimulation}
% \end{figure}
% %%%%%%%%%%%%%%%%%%%%%%%%%%%%%%%%%%%%%%%%%%%%%%%%%%%%%%%%
\begin{table}[t]
\centering
    \ra{1.2}
    \caption{Mean absolute errors in simulation}
    \begin{tabular}{@{\extracolsep{\fill}}l|ccc@{}}
    \hline
    Simulated value & $f$ & LSTM & NN \\ \hline 
    $x_t$ ($1 \times 10^{-4} m$) & \textbf{0.34} & 3.86 & 9.86 \\ 
    %$u_t$ ($1 \times 10^{-4} m/s$) & 8 & A & A \\ 
    F$_{x,t}$ ($1 \times 10^{-4} N$) & 3.68 & $-$ & $-$ \\ 
    F$_{y,t}$ ($1 \times 10^{-4} N$) & 12.01 & $-$ & $-$ \\ \hline
   \end{tabular}
    \label{tab:openloopsimulation}
\end{table}
%%%%%%%%%%%%%%%%%%%%%%%%%%%%%%%%%%%%%%%%%%%%%%%%%%%%%%%%

%%%%%%%%%%%%%%%%%%%%%%%%%%%%%%%%%%%%%%%%%%%%%%%%%%%%%%%%%%%%%%%%%%%%%%%%%%%%%%%%

\begin{figure}[t!]
    \centering
    \includegraphics[width=0.4\textwidth]{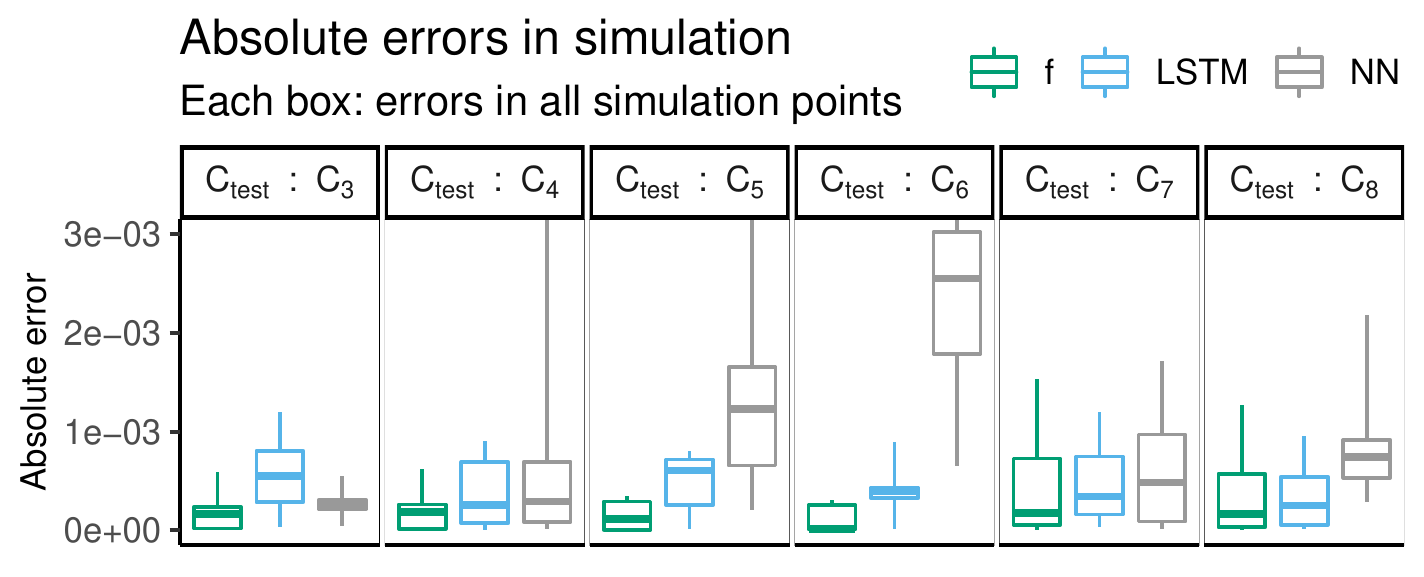}
    \caption{Absolute error in simulated location $x_t$. Each box point has data for all simulation points.}
    \label{fig:openlooperros}
\end{figure}

%\oldtext{To evaluate our physics-informed model as a simulator for new setups (defined by parameters of system dynamics, \qref{q:simulation}), we study the absolute errors in simulating the data across test setups.}
% \Fref{fig:openloopsimulation} shows the location ($x_t$) and forces ($F_y$) simulated using physics-informed model $f$, a baseline models, and the FEM model. 
\tref{tab:openloopsimulation} provides the mean absolute errors in simulating location ($x_t$) and forces using physics-informed model $f$ and the baseline models across all test setups. \Fref{fig:openlooperros} shows the absolute error in $x$ for all test setups for $f$, $N\!N$, and $L\!S\!T\!M$. In all cases, $f$ outperforms the baseline models. Additionally, the average absolute error in $x$ simulated using $f$ is less than 0.05 times the magnitude of the actual values, \ie we note approximately~$\le$\,5\% error compared the FEM simulation. 
%We present the location ($x_t$) and forces applied by the DEA on the top surface of the coin ($F_x$ and $F_y$) of the coin. 
%The solid lines present the data generated by the FEM model.

In the first half of the simulation time during static friction (\sref{sec:SetupD}), we see negligible displacements and increasing $F_{y}$. In the latter half, the kinetic frictional force becomes stable, and we see a change in coin location ({$x_t$}). Our model $f$ accurately estimates $x$ in both stages. % During kinetic friction, $f$ outperforms the baseline $N\!N$ and $L\!S\!T\!M$. %\oldtext{where the location simulated by $N\!N$ increases exponentially.}
% \Fref{fig:openlooperros} shows the absolute error in $x$ for all test setups for $f$, $N\!N$, and $L\!S\!T\!M$. In all cases, $f$ outperforms the baseline models. Additionally, the average absolute error in $x$ simulated using $f$ is less than 0.05 times the magnitude of the actual values, \ie we note approximately~$\le$\,5\% error compared the FEM simulation. 
We see a similar accuracy for $f$ compared to the FEM, $N\!N$, and $L\!S\!T\!M$ in simulating velocity $u_x$ for the coin. 
\newtext{While decreasing the neurons per layer in $f$ from 128 to 64 (\sref{sec:parametersettging}) results in no statistically significant increase in average error, decreasing the capacity of baseline $N\!N$ by the same amount results in an increase from 15\% average error to 50\%.}

\newtext{Real-time control in manipulation tasks requires fast simulations. For 1 second of simulation with 700 points, the FEM model takes 130 seconds (average for all test coins) --- a prohibitive duration for real-time control. In comparison, our physics-informed model $f$ takes $\le$ 0.7 seconds. While both baseline models are also faster than FEM, (0.3 and 1.2 seconds for $N\!N$ and $L\!S\!T\!M$ respectively), they provide poor accuracy compared to $f$. 
% We further note that similar accuracy of $f$ for a less complex architecture (\ie less number of neurons) for material network $m$, whereas for the baseline $N\!N$, accuracy decreases with less complex architectures.  
Our proposed model $f$ is fast as opposed to FEM but provides high fidelity to FEM as a simulation model of soft robotic manipulation.}

The absolute error in forces $F_x$ and $F_y$ in the region of static friction is higher compared to the region of kinetic friction. This happens because we optimize material network $m$ using a physics informed loss function, \ie a loss function that is based on the error in the next state $s$ and the error in the interaction variables $z$ (\eref{eqn:loss_learning}). Optimizing $m$ using this loss function assists in learning the overall manipulation task behaviour, as opposed to only learning the outputs of $m$ ($F_x$ and $F_y$).
This behaviour is non-restrictive, as the objective of our model is to learn the next state of the motion, which is simulated accurately. 

%Additionally, the simulated value of force in $x$-direction is lower than the FEM generated value. This happens because the dynamics of our physics based models does not consider the friction on the top surface ($\mu_b$), which is be present in the realistic scenario (and in FEM model). 

%Once trained, model $f$ can simulate data for new setups by changing dynamics parameters (like mass $m_c$). This helps in answering~\qref{q:simulation} (simulation accuracy of $f$ for system with different parameter setting?), as we can conclude from~\fref{fig:openlooperros} that our physics-informed model has $\le$\,5\% error compared to FEM and outperforms the baseline networks.

%%%%%%%%%%%%%%%%%%%%%%%%%%%%%%%%%%%%%%%%%%%%%%%%%%%%%%%%%%%%%%%%%%%%%%%%%%%%%%%%
%%%%%%%%%%%%%%%%%%%%%%%%%%%%%%%%%%%%%%%%%%%%%%%%%%%%%%%%%%%%%%%%%%%%%%%%%%%%%%%%

%%%%%%%%%%%%%%%%%%%%%%%%%%%%%%%%%%%%%%%%%%%%%%%%%%%%%%%%%%%%%%%%%%%%%%%%%%%%%%%%
\subsection{Control (\qref{q:control})}
\label{sec:controlresults}
%%%%%%%%%%%%%%%%%%%%%%%%%%%%%%%%%%%%%%%%%%%%%%%%%%%%%%%%%%%%%%%%%%%%%%%%%%%%%%%%

%%%%%%%%%%%%%%%%%%%%%%%%%%%%%%%%%%%%%%%%%%%%%%%%%%%%%
%\paragraph{\begin{experiment}\label{exp:controlexpt} Control and inference\end{experiment}}
%%%%%%%%%%%%%%%%%%%%%%%%%%%%%%%%%%%%%%%%%%%%%%%%%%%%%
In this experiment, we evaluate $f$ to answer~\qref{q:control} (performance of model-based MPC compared to other control policies?).
In the control step (\algoref{Algo:control}), we learn closed-loop control for test setups $C_1$ and $C_2$.
Prior to this, in the learning step (\algoref{Algo:learning}), we train the model $f_{a}$ using the FEM data from setups $\{C_3,\dots, C_8\}$.
We do not use the data from $C_1$ and $C_2$ during learning to avoid information leakage.  
%For inference, $m_c$ is set to 0.001 $g$ in $f_a$. We use~ to infer $m_c$ and take control actions in the closed-loop setting. Timestep $\Delta t$ is set to 0.001 $s$. 
%(model-based control policy
%(based on f), compared to other control policies?)
%MPC controller is compared to a baseline controller, that uses a naive policy. The controller increases the voltage after each iteration (\ie voltage increases by 1 $V$ after each iteration). 
For $C_1$ and $C_2$, we learn the following policies:
\begin{enumerate}[(i)]
\item \textit{MPC policy}: A model-based control policy defined using differentiable model $f$ and an MPC solver~\cite{Brandon_2019_MPCDiff}. \newtext{The MPC uses the model with infered physical parameters.}
\item \textit{SAC policy}: A model-free Actor-Critic policy learnt using the Soft Actor-Critic algorithm~\cite{haarnoja_2018_SAC},
\item \textit{PD policy}: A feedback based control policy (previously tested for DEA control~\cite{Karner_2021_DEAPIDcontrol}),
\item \textit{Heur policy}: A heuristic control policy that linearly ramps up actuation voltage (\ie voltage increases by 0.5\,V after each iteration until terminal state). 
\end{enumerate}

The \textit{Heur policy} is inspired by typical soft-robotic control policies~\cite{Yin_2021_ModellingDeformable}.
The environment is simulated by trained physics-informed models $f_{s,1}$ and $f_{s,2}$. This is due to the lack of a real-world DEA setup (However, \sref{sec:simresults} shows our physics-informed models are accurate simulators). 
%We report coin location and inferred $m_c$.

%%%%%%%%%%%%%%%%%%%%%%%%%%%%%%%%%%%%%%%%%%%%%%%%%%%%%%%%%%%%%%%%%%%%%%%%%%%%%%%%
\begin{figure}[t!]
    \centering
    \includegraphics[width=0.4\textwidth]{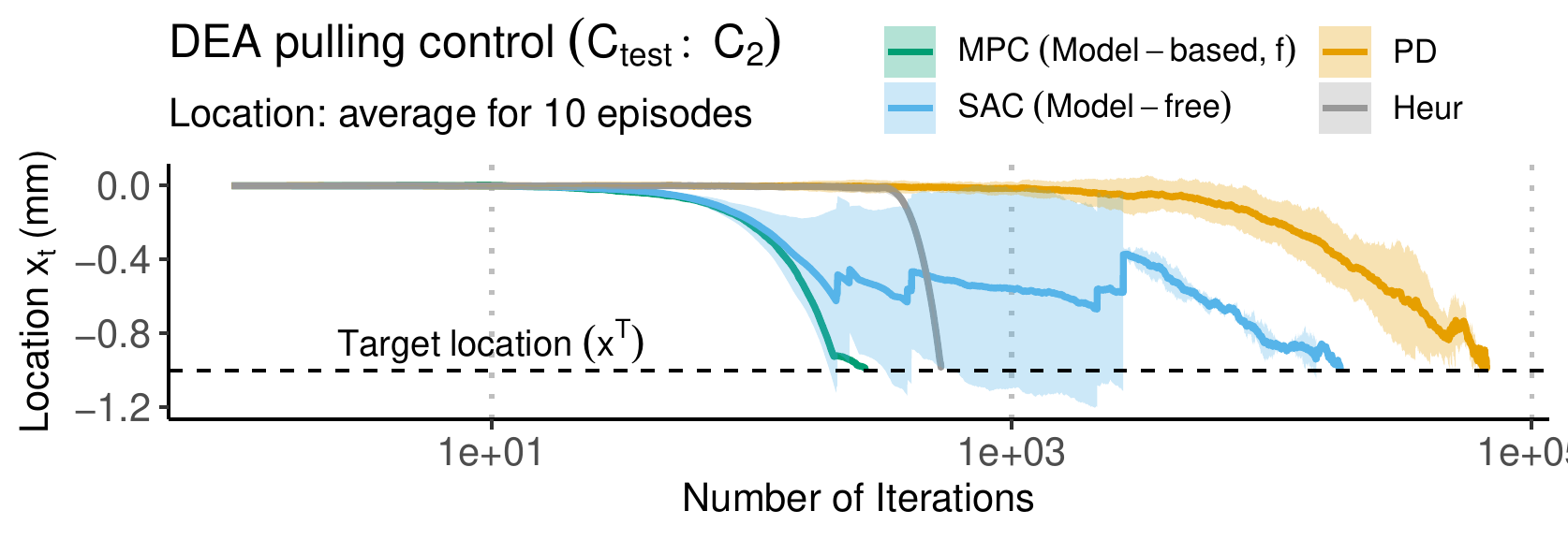}
    \caption{DEA pulling control for $C_2$ . Terminal state: coin is $\le$ 0.01 mm from $x^T$ (Solid line: average for 10 `episodes', shaded region: 25-75\%)}
    \label{fig:closeloopcontrol}
\end{figure}
%%%%%%%%%%%%%%%%%%%%%%%%%%%%%%%%%%%%%%%%%%%%%%%%%%%%%%%%%%%%%%%%%%%%%%%%%%%%%%%%

%\oldtext{This section presents results for~\expref{exp:controlexpt}, which aims to evaluate closed-loop control (\qref{q:control}) and inference (\qref{q:inference}) defined in the proposed framework. In DEA pulling, using~\algoref{Algo:control}, we learn a model-based control policy by utilizing the differentiable $f$ and an MPC solver.}
%A control `episode' terminates when the coin reaches $le$ 0.01 $mm$ from the target location.
%with achieveing  estimated from the differentialble physics based model provides the action sequences, \ie actuation voltage of the DEA.
%This voltage causes change in the location of coin. Controller tries to reach the target state while penalizing the actions. 
\Fref{fig:closeloopcontrol} shows the average coin location $x_t$ during closed-loop control of test setup $C_2$. Average is calculated across 10 episodes to compare the~\textit{MPC policy} (model-based), with~\textit{SAC policy} (model-free), a~\textit{PD policy} and a~\textit{Heur policy}. We notice similar results for both test setups $C_1$ and $C_2$. 
\newtext{The variance in coin position is due to the noise added in the simulated observations to estimate a realistic scenario.}

\newtext{The coin reaches the target in $\le$\,200 iterations under the \textit{MPC policy}, \ie $\le$\,200 observations are captured in the control loop. In contrast, \textit{SAC policy}, \textit{PD policy}, and \textit{Heur} capture approximately 10,000, 1500, and 500 iterations. Thus, during control, the \textit{MPC policy} reaches the target in the least number of steps, where the computation time per iteration (average 1.2\,s) is not computationally prohibitive for a real world manipulation. We note similar computation times for the model-free \textit{SAC policy} (average 0.005\,s) and \textit{PD policy} (average 0.001\,s).}
%Due to the gradient-based optimization, \textit{MPC policy} takes longer per iteration (1.2\,s) compared to the model-free \textit{SAC policy} (0.005\,s), \textit{PD policy} (0.001\,s), and \textit{Heur policy} (0.002\,s). \textit{MPC policy} has lowest data storage requirements and, once computed, takes least steps to reach the target.}
%\to reach the target. We further note that the \textit{MPC policy} outperforms the

During control using \textit{Heur policy} we notice sudden motion towards $x^T$ after approximately 280 iterations. This represents the transition from stage of static friction to stage of kinetic friction. 
The \textit{Heur policy} linearly ramps up actuation voltage every iteration, and thus, does not depend on location feedback. 
In contrast, both \textit{PD policy} and \textit{SAC policy} rely on observed location, and take longer to reach and manipulate the coin in the stage of kinetic friction. 
%Additionally, MPC converges to $\le$ 0.001 $mm$ from the target, compared to $\le$ 0.01 $mm$ for the baseline. The MPC controller is 70\% faster in reaching the target compared to the baseline. 

%\oldtext{We answered \qref{q:control} and \qref{q:inference} in this section, and conclude that our framework learns accurate physics-informed model $f$ that can be used as a simulator for inference and developing closed-loop model-based control policies. In control, a model-based~\textit{MPC policy} outperformed all other policies with an order of hundreds of iterations, and we note $\le$\,10\% in parameter inference.}

% %%%%%%%%%%%%%%%%%%%%%%%%%%%%%%%%%%%%%%%%%%%%%%%%%%%%%%%%%%%%%%%%%%%%%%%%%%%%%%%%
% \begin{figure}[t!]
%     \centering
%     \includegraphics[width=0.4\textwidth]{Figures/Inference-of-parameters.pdf}
%     \caption{Inference of mass of the coin and frictional coefficient by DEA pulling. (Reported after every 256 iterations to keep figure legible) }
%     \label{fig:massinference}
% \end{figure}
% %%%%%%%%%%%%%%%%%%%%%%%%%%%%%%%%%%%%%%%%%%%%%%%%%%%%%%%%%%%%%%%%%%%%%%%%%%%%%%%%

%%%%%%%%%%%%%%%%%%%%%%%%%%%%%%%%%%%%%%%%%%%%%%%%%%%%%%%%%%%%%%%%%%%%%%%%%%%%%%%%
\subsection{Inference (\qref{q:inference})}
\label{sec:inferenceresults}
%%%%%%%%%%%%%%%%%%%%%%%%%%%%%%%%%%%%%%%%%%%%%%%%%%%%%%%%%%%%%%%%%%%%%%%%%%%%%%%%

\newtext{This experiment evaluates how accurately $f$ infers the model parameters. \algoref{Algo:control} infers $m_c$ and $\mu_b$ for test setups $C_1$ and $C_2$. Similar to the control experiment, in the learning step, we train $f_{a}$ using FEM data from setups $\{C_3,\dots, C_8\}$.} % \Fref{fig:massinference} shows 
The mass $m_c$ and frictional coefficient $\mu_b$ inferred by physics-informed model $f$ are compared against their real values (used in $C_\textit{test}$ during control with \textit{MPC policy}). Inferred $m_c$ converges to $\le$\,10\% error compared to the real value within 2,000 iterations, while $\mu_b$, the converges within 300 iterations. Similar results hold across both test setups and all episodes. 

%%%%%%%%%%%%%%%%%%%%%%%%%%%%%%%%%%%%%%%%%%%%%%%%%%%%%%%%%%%%%%%%%%%%%%%%%%%%%%%%
%%%%%%%%%%%%%%%%%%%%%%%%%%%%%%%%%%%%%%%%%%%%%%%%%%%%%%%%%%%%%%%%%%%%%%%%%%%%%%%%
\section{Conclusions}
\label{sec:conclusions}
%%%%%%%%%%%%%%%%%%%%%%%%%%%%%%%%%%%%%%%%%%%%%%%%%%%%%%%%%%%%%%%%%%%%%%%%%%%%%%%%
%%%%%%%%%%%%%%%%%%%%%%%%%%%%%%%%%%%%%%%%%%%%%%%%%%%%%%%%%%%%%%%%%%%%%%%%%%%%%%%%

This paper presents a framework to learn a differentiable simulator to develop a corresponding controller for soft robotic manipulation. We defined a physics-informed model $f$ consisting of a material network $m$, and dynamics $d$. This model $f$ can be used as a simulator for data-generation, inference, and control policy optimization. We designed a soft-robotics case study where a coin is pulled using unimorph DEA. FEM simulation of the DEA generated data to train $f$. Our experiments used multiple setups to evaluate the framework in learning $f$ and model-based control.

From our analyses, we conclude,
\begin{enumerate*}[(i)]
    \item the physics-informed model $f$ trained using the proposed framework can simulate new setups (characterized by parameters $\phi$) with $\le 5\%$ error compared to FEM (\fref{fig:openlooperros}); 
    %The loss function ~\eref{eqn:loss_learning} assists in learning overall task behaviour as opposed to merely learning the deformable material behaviour in $m$ (\sref{sec:simresults}).
    \item a closed-loop \textit{MPC policy} based on differentiable model $f$ outperformed all other policies in orders of hundreds of iterations (\sref{sec:controlresults}); 
    \item $f$ can be used for accurate inference of the parameters $\phi$: $m_c$ and $\mu_b$ (\sref{sec:inferenceresults}). 
\end{enumerate*}
Future research directions include: \newtext{Further evaluating this control framework with \emph{physical} soft robotic actuators}; \newtext{Evaluating the model $f$ in tasks with large degrees of freedom}; and exploring model-based policies with lower computational requirements compared to MPC.
%%%%%%%%%%%%%%%%%%%%%%%%%%%%%%%%%%%%%%%%%%%%%%%%%%%%%%%%%%%%%%%%%%%%%%%%%%%%%%%%
%%%%%%%%%%%%%%%%%%%%%%%%%%%%%%%%%%%%%%%%%%%%%%%%%%%%%%%%%%%%%%%%%%%%%%%%%%%%%%%%
%%%%%%%%%%%%%%%%%%%%%%%%%%%%%%%%%%%%%%%%%%%%%%%%%%%%%%%%%%%%%%%%%%%%%%%%%%%%%%%%
% \section*{APPENDIX}
%%%%%%%%%%%%%%%%%%%%%%%%%%%%%%%%%%%%%%%%%%%%%%%%%%%%%%%%%%%%%%%%%%%%%%%%%%%%%%%%
%%%%%%%%%%%%%%%%%%%%%%%%%%%%%%%%%%%%%%%%%%%%%%%%%%%%%%%%%%%%%%%%%%%%%%%%%%%%%%%%

%%%%%%%%%%%%%%%%%%%%%%%%%%%%%%%%%%%%%%%%%%%%%%%%%%%%%%%%%%%%%%%%%%%%%%%%%%%%%%%%
%%%%%%%%%%%%%%%%%%%%%%%%%%%%%%%%%%%%%%%%%%%%%%%%%%%%%%%%%%%%%%%%%%%%%%%%%%%%%%%%
\section*{ACKNOWLEDGMENT}
S. Ramamoorthy would like to acknowledge financial support from the Alan Turing Institute, for the project `Enabling advanced autonomy through human-AI collaboration'. We thank Mukul Sahu for his comments on FEM modeling.
%%%%%%%%%%%%%%%%%%%%%%%%%%%%%%%%%%%%%%%%%%%%%%%%%%%%%%%%%%%%%%%%%%%%%%%%%%%%%%%%
%%%%%%%%%%%%%%%%%%%%%%%%%%%%%%%%%%%%%%%%%%%%%%%%%%%%%%%%%%%%%%%%%%%%%%%%%%%%%%%%

%%%%%%%%%%%%%%%%%%%%%%%%%%%%%%%%%%%%%%%%%%%%%%%%%%%%%%%%%%%%%%%%%%%%%%%%%%%%%%%%
%%%%%%%%%%%%%%%%%%%%%%%%%%%%%%%%%%%%%%%%%%%%%%%%%%%%%%%%%%%%%%%%%%%%%%%%%%%%%%%%
% \section*{SUPPLEMENTARY MATERIALS}
% We have made videos of the manipulation task and code available
% - see details at:
% \href{https://sites.google.com/view/phy-informed-sim-soft-robot/home}{https://sites.google.com/view/phy-informed-sim-soft-robot/home}

%%%%%%%%%%%%%%%%%%%%%%%%%%%%%%%%%%%%%%%%%%%%%%%%%%%%%%%%%%%%%%%%%%%%%%%%%%%%%%%%
%%%%%%%%%%%%%%%%%%%%%%%%%%%%%%%%%%%%%%%%%%%%%%%%%%%%%%%%%%%%%%%%%%%%%%%%%%%%%%%%
%%%%%%%%%%%%%%%%%%%%%%%%%%%%%%%%%%%%%%%%%%%%%%%%%%%%%%%%%%%%%%%%%%%%%%%%%%%%%%%%
\addtolength{\textheight}{-12cm}   
%%%%%%%%%%%%%%%%%%%%%%%%%%%%%%%%%%%%%%%%%%%%%%%%%%%%%%%%%%%%%%%%%%%%%%%%%%%%%%%%
\bibliographystyle{IEEEtran}
\bibliography{main.bib}

\begin{thebibliography}{10}
\providecommand{\url}[1]{#1}
\csname url@rmstyle\endcsname
\providecommand{\newblock}{\relax}
\providecommand{\bibinfo}[2]{#2}
\providecommand\BIBentrySTDinterwordspacing{\spaceskip=0pt\relax}
\providecommand\BIBentryALTinterwordstretchfactor{4}
\providecommand\BIBentryALTinterwordspacing{\spaceskip=\fontdimen2\font plus
\BIBentryALTinterwordstretchfactor\fontdimen3\font minus
  \fontdimen4\font\relax}
\providecommand\BIBforeignlanguage[2]{{%
\expandafter\ifx\csname l@#1\endcsname\relax
\typeout{** WARNING: IEEEtran.bst: No hyphenation pattern has been}%
\typeout{** loaded for the language `#1'. Using the pattern for}%
\typeout{** the default language instead.}%
\else
\language=\csname l@#1\endcsname
\fi
#2}}

\bibitem{KIM_2013_softrobots}
S.~Kim, C.~Laschi, and B.~Trimmer, ``Soft robotics: a bioinspired evolution in
  robotics,'' \emph{Trends in Biotechnology}, vol.~31, 2013.

\bibitem{Rus_2015_softrobotics}
D.~Rus and M.~Tolley, ``Design, fabrication and control of soft robots,''
  \emph{Nature}, vol. 521, 2015.

\bibitem{Yin_2021_ModellingDeformable}
H.~Yin, A.~Varava, and D.~Kragic, ``Modeling, learning, perception, and control
  methods for deformable object manipulation,'' \emph{Science Robotics},
  vol.~6, 2021.

\bibitem{Atab_2020_SoftReview}
N.~El-Atab, R.~Mishra, F.~Al-modaf, L.~Joharji, A.~Alsharif, H.~Alamoudi,
  M.~Diaz, N.~Qaiser, and M.~Mustafa, ``Soft actuators for soft robotic
  applications: A review,'' \emph{Advanced Intelligent Systems}, vol.~2, 2020.

\bibitem{Todorov_2012_mujoco}
E.~Todorov, T.~Erez, and Y.~Tassa, ``Mujoco: A physics engine for model-based
  control,'' in \emph{Proceedings of the IEEE/RSJ International Conference on
  Intelligent Robots and Systems (IROS)}, 2012, pp. 5026--5033.

\bibitem{Brandon_2019_MPCDiff}
B.~Amos, I.~D.~J. Rodriguez, J.~Sacks, B.~Boots, and J.~Z. Kolter,
  ``Differentiable {MPC} for end-to-end planning and control,'' in
  \emph{Proceedings of the 32nd International Conference on Neural Information
  Processing Systems (NIPS)}, 2018, p. 8299–8310.

\bibitem{haarnoja_2018_SAC}
T.~Haarnoja, A.~Zhou, P.~Abbeel, and S.~Levine, ``Soft actor-critic: Off-policy
  maximum entropy deep reinforcement learning with a stochastic actor,'' in
  \emph{Proceedings of the 35th International Conference on Machine Learning
  (ICML)}, 2018, pp. 1861--1870.

\bibitem{Karner_2021_DEAPIDcontrol}
T.~Karner and J.~Gotlih, ``Position control of the dielectric elastomer
  actuator based on fractional derivatives in modelling and control,''
  \emph{Actuators}, vol.~10, 2021.

\bibitem{tataia_2017_pouring}
T.~Lopez-Guevara, N.~K. Taylor, M.~U. Gutmann, S.~Ramamoorthy, and K.~Subr,
  ``Adaptable pouring: Teaching robots not to spill using fast but approximate
  fluid simulation,'' in \emph{Proceedings of the 1st Annual Conference on
  Robot Learning}, 2017, pp. 77--86.

\bibitem{Coevoet_2019_cloth}
E.~Coevoet, A.~Escande, and C.~Duriez, ``Soft robots locomotion and
  manipulation control using fem simulation and quadratic programming,'' in
  \emph{Proceedings of 2nd IEEE International Conference on Soft Robotics
  (RoboSoft)}, 2019, pp. 739--745.

\bibitem{Muzel_2020_FEMcomposite}
S.~David~Müzel, E.~Bonhin, N.~Guimarães, and E.~Guidi, ``Application of the
  finite element method in the analysis of composite materials: A review,''
  \emph{Polymers}, vol.~12, 2020.

\bibitem{Gupta_2019_DEAreview}
U.~Gupta, L.~Qin, Y.~Wang, H.~Godaba, and J.~Zhu, ``Soft robots based on
  dielectric elastomer actuators: a review,'' \emph{Smart Materials and
  Structures}, vol.~28, 2019.

\bibitem{Zhou_2020_DEAFEM}
F.~Zhou, X.~Yang, Y.~Xiao, Z.~Zhu, T.~Li, and Z.~Xu, ``Electromechanical
  analysis and simplified modeling of dielectric elastomer multilayer bending
  actuator,'' \emph{AIP Advances}, vol.~10, 2020.

\bibitem{Shintake_2016_DEAUnderwater}
J.~Shintake, H.~Shea, and D.~Floreano, ``Biomimetic underwater robots based on
  dielectric elastomer actuators,'' in \emph{Proceedings of IEEE/RSJ
  International Conference on Intelligent Robots and Systems (IROS)}, 2016, pp.
  4957--4962.

\bibitem{DuDuta_2020_multimodel_locomotion}
M.~Duduta, F.~Berlinger, R.~Nagpal, D.~R. Clarke, R.~J. Wood, and F.~Z. Temel,
  ``Tunable multi-modal locomotion in soft dielectric elastomer robots,''
  \emph{IEEE Robotics and Automation Letters}, vol.~5, 2020.

\bibitem{Cao_2018_DEAmodeling}
J.~Cao, W.~Liang, Q.~Ren, U.~Gupta, F.~Chen, and J.~Zhu, ``Modelling and
  control of a novel soft crawling robot based on a dielectric elastomer
  actuator,'' in \emph{2018 IEEE International Conference on Robotics and
  Automation (ICRA)}, 2018, pp. 4188--4193.

\bibitem{Araromi_2012_FEMDEA}
O.~Araromi and S.~Burgess, ``\BIBforeignlanguage{English}{A finite element
  approach for modelling multilayer unimorph dielectric elastomer actuators
  with inhomogeneous layer geometry},''
  \emph{\BIBforeignlanguage{English}{Smart Materials and Structures}}, vol.~21,
  2012.

\bibitem{Zhao_2008_FEMDE}
X.~Zhao and Z.~Suo, ``Method to analyze programmable deformation of dielectric
  elastomer layers,'' \emph{Applied Physics Letters}, vol.~93, 2008.

\bibitem{Prajjwal_2019_cutting}
P.~Jamdagni and Y.-B. Jia, ``Robotic cutting of solids based on fracture
  mechanics and fem,'' in \emph{2019 IEEE/RSJ International Conference on
  Intelligent Robots and Systems (IROS)}, 2019, pp. 8252--8257.

\bibitem{Heiden_2021_DiSECt}
E.~Heiden, M.~Macklin, Y.~S. Narang, D.~Fox, A.~Garg, and F.~Ramos, ``{DiSECt:
  A Differentiable Simulation Engine for Autonomous Robotic Cutting},'' in
  \emph{Proceedings of Robotics: Science and Systems}, 2021.

\bibitem{Bern_2020_controlmodel}
J.~M. Bern, Y.~Schnider, P.~Banzet, N.~Kumar, and S.~Coros, ``Soft robot
  control with a learned differentiable model,'' in \emph{2020 3rd IEEE
  International Conference on Soft Robotics (RoboSoft)}, 2020, pp. 417--423.

\bibitem{George_2019_modelbasedcontrol}
T.~G. Thuruthel, E.~Falotico, F.~Renda, and C.~Laschi, ``Model-based
  reinforcement learning for closed-loop dynamic control of soft robotic
  manipulators,'' \emph{IEEE Transactions on Robotics}, vol.~35, no.~1, pp.
  124--134, 2019.

\bibitem{raissi_2019_pinn}
M.~Raissi, P.~Perdikaris, and G.~E. Karniadakis, ``Physics-informed neural
  networks: A deep learning framework for solving forward and inverse problems
  involving nonlinear partial differential equations,'' \emph{Journal of
  Computational Physics}, vol. 378, 2019.

\bibitem{Pelrine_1998_DEA}
R.~E. Pelrine, R.~D. Kornbluh, and J.~P. Joseph, ``Electrostriction of polymer
  dielectrics with compliant electrodes as a means of actuation,''
  \emph{Sensors and Actuators A: Physical}, vol.~64, 1998.

\bibitem{Smith_2009_abaqus}
M.~Smith, \emph{\BIBforeignlanguage{English}{ABAQUS/Standard User's Manual,
  Version 6.9}}.\hskip 1em plus 0.5em minus 0.4em\relax United States: Dassault
  Syst{\`e}mes Simulia Corp, 2009.

\bibitem{Zhang_2019_DEAnonlinear}
H.~Zhang, M.~Dai, and Z.~Zhang, ``Application of viscoelasticity to nonlinear
  analyses of circular and spherical dielectric elastomers,'' \emph{AIP
  Advances}, vol.~9, no.~4, p. 045010, 2019.

\bibitem{Piefort_2000_FEMPeiso}
V.~Piefort, ``Finite element modeling of piezoelectric structures,'' 2000.

\bibitem{Wang_2017_DEProp}
N.~Wang, C.~Chaoyu, H.~Guo, B.~Chen, and X.~Zhang, ``Advances in dielectric
  elastomer actuation technology,'' \emph{Science China Technological
  Sciences}, vol.~61, 2017.

\bibitem{paszke_2017_pytorch}
A.~Paszke, S.~Gross, S.~Chintala, G.~Chanan, E.~Yang, Z.~DeVito, Z.~Lin,
  A.~Desmaison, L.~Antiga, and A.~Lerer, ``Automatic differentiation in
  {PyTorch},'' 2017.

\end{thebibliography}
%%%%%%%%%%%%%%%%%%%%%%%%%%%%%%%%%%%%%%%%%%%%%%%%%%%%%%%%%%%%%%%%%%%%%%%%%%%%%%%%
\end{document}